\documentclass{article}
\usepackage{arxiv}

\usepackage[utf8]{inputenc} 
\usepackage[T1]{fontenc}    
\usepackage[pagebackref=true,
    breaklinks=true,
    colorlinks,
    linkcolor=blue,
    citecolor=blue,
    filecolor=black,
    urlcolor=blue,
    bookmarks=false]{hyperref}
\usepackage{url}            
\usepackage{booktabs}       
\usepackage{amsfonts}       
\usepackage{nicefrac}       
\usepackage{graphicx}
\usepackage{amssymb}
\usepackage{multirow}
\usepackage{caption}
\usepackage{subcaption}

\usepackage{listings}
\usepackage{xcolor}

\definecolor{codegreen}{rgb}{0,0.6,0}
\definecolor{codegray}{rgb}{0.5,0.5,0.5}
\definecolor{codepurple}{rgb}{0.58,0,0.82}
\definecolor{backcolour}{rgb}{0.95,0.95,0.92}

\lstdefinestyle{mystyle}{
    backgroundcolor=\color{backcolour},
    commentstyle=\color{codegreen},
    keywordstyle=\color{magenta},
    numberstyle=\tiny\color{codegray},
    stringstyle=\color{codepurple},
    basicstyle=\ttfamily\footnotesize,
    breakatwhitespace=false,
    breaklines=true,
    captionpos=b,
    keepspaces=true,
    numbers=left,
    numbersep=5pt,
    showspaces=false,
    showstringspaces=false,
    showtabs=false,
    tabsize=2
}

\usepackage{amsmath}
\usepackage{amsthm}
\usepackage{fontawesome}
\usepackage{makecell}

\usepackage{color}
\usepackage{hyperref}
\usepackage{longtable}

\newtheorem{theorem}{Theorem}[section]

\newtheorem{example}{Example}

\graphicspath{ {./images/} }

\def\[#1\]{$$#1$$}

\def\frac#1#2{{#1\over#2}}

\newcommand{\draft}[1]{{#1}}

\newcommand{\sm}[1]{{#1}}

\newcommand{\paride}[1]{{\textsc{#1}}}
\newcommand{\paridelist}[1]{{\textsc{#1}}}

\renewcommand{\arraystretch}{1.5}

\title{Logic Explained Networks}

\author{Gabriele Ciravegna\thanks{Equal contribution}\hspace{1.2mm}$^{1,2}$\hspace{1.2mm}\href{https://orcid.org/0000-0002-6799-1043}{\includegraphics[scale=0.06]{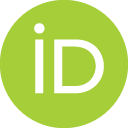}},
Pietro Barbiero$^{*3}$\hspace{1mm}\href{https://orcid.org/0000-0003-3155-2564}{\includegraphics[scale=0.06]{orcid.png}},
Francesco Giannini$^{*2}$, \\
\textbf{
Marco Gori$^{2,4}$\hspace{1mm}\href{https://orcid.org/0000-0001-6337-5430}{\includegraphics[scale=0.06]{orcid.png}},
Pietro Li\'o$^{3}$,
Marco Maggini$^{2}$,
Stefano Melacci$^{2}$\hspace{1mm}\href{https://orcid.org/0000-0002-0415-0888}{\includegraphics[scale=0.06]{orcid.png}}
}\\
$^{1}$Università di Firenze (Italy), $^{2}$Università di Siena (Italy), \\ $^{3}$University of Cambridge (UK), $^{4}$Universitè Côte d’Azur (France)\\
\texttt{gabriele.ciravegna@unifi.it, pb737@cam.ac.uk, francesco.giannini@unisi.it}\\
\texttt{marco.gori@unisi.it, pl219@cam.ac.uk, marco.maggini@unisi.it, mela@diism.unisi.it}
}

\begin{document}
\maketitle

\begin{abstract}
The large and still increasing popularity of deep learning clashes with a major limit of neural network architectures, that consists in their lack of capability in providing human-understandable motivations of their decisions.
In situations in which the machine is expected to support the decision of human experts, providing a comprehensible explanation is a feature of crucial importance.
The language used to communicate the explanations must be formal enough to be implementable in a machine and friendly enough to be understandable by a wide audience.
In this paper, we propose a general approach to Explainable Artificial Intelligence in the case of neural architectures, showing how a mindful design of the networks leads to a family of interpretable deep learning models called Logic Explained Networks (LENs).
LENs only require their inputs to be 
human-understandable predicates, and they provide explanations in terms of simple First-Order Logic (FOL) formulas involving such predicates. LENs are general enough to cover a large number of scenarios. Amongst them, we consider the case in which LENs are directly used as special classifiers with the capability of being explainable, or when they act as additional networks with the role of creating the conditions for making a black-box classifier explainable by FOL formulas.
Despite supervised learning problems are mostly emphasized,
we also show that LENs can learn and provide explanations in unsupervised learning settings.
Experimental results on several datasets and tasks show that LENs may yield better classifications than established white-box models, such as decision trees and Bayesian rule lists, while providing more compact and meaningful explanations.
\end{abstract}

\section{Introduction}
The application of deep neural networks in safety-critical domains has been strongly limited \cite{chander2018working}, since neural networks are generally considered black-boxes whose decision processes are opaque or too complex to be  understood by users. 
Employing black-box\footnote{In the context of this paper, a black-box classifier is any classifier that cannot provide human understandable explanations about its decision.} models may be unacceptable in contexts such as industry, medicine or courts, where the potential economical or ethical repercussions are calling for lawmakers to discourage from a reckless application of non-interpretable models \cite{gdpr2017,law10code,goddard2017eu,gunning2017explainable}. As a consequence, research in Explainable Artificial Intelligence (XAI) has become strategic and has been massively encouraged, leading to the development of a variety of techniques that aim at explaining black-box models \cite{das2020opportunities,brundage2020toward} or at designing interpretable models \cite{carvalho2019machine,rudin2019stop}.

The notions of \emph{interpretability} and \emph{explainability} were historically used as synonyms. However the distinction between interpretable models and models providing explanations has now become more evident, as recently discussed by different authors \cite{gilpin2018explaining,lipton2018mythos,marcinkevivcs2020interpretability}.
Even if there are no common accepted formal definitions, a model is considered interpretable when its decision process is generally transparent and can be understood directly by its structure and parameters, such as \emph{linear models} or \emph{decision trees}. On the other hand, the way an existing (black-box) model makes predictions can be explained by a surrogate interpretable model or by means of techniques providing intelligible descriptions of the model behaviour by e.g. formal rules, saliency maps, question-answering. 
In some contexts, the use of a black-box model may be unnecessary or even not preferable \cite{doshi2017towards,doshi2018considerations,ahmad2018interpretable,rudin2019stop,samek2020toward,rudin2021interpretable}. For instance, a proof of concept, a prototype, or the solution to a simple classification problem can be easily based on standard interpretable by design 
AI solutions \cite{breiman1984classification, schmidt2009distilling, letham2015interpretable, cranmer2019learning, molnar2020interpretable}.
However, interpretable models may generally miss to capture complex relationships among data. Hence, in order to achieve state-of-the-art performance in more challenging problems, it may be necessary to leverage black-box models \cite{battaglia2018relational,devlin2018bert,dosovitskiy2020image,xie2020self} that, in turn, may require an additional explanatory model to gain the trust of the user.

In the literature, there is a wide consensus on the necessity of having an explanation for machine learning models that are employed in safety-critical domains, while there is not even agreement on what an \emph{explanation} actually is, nor if it could be formally defined \cite{doshi2017towards,lipton2018mythos}.
As observed by Srinivasan and Chander, explanations should serve cognitive-behavioral purposes such as engendering trust, aiding bias identification, or taking actions/decisions \cite{srinivasanexplanation}. An explanation may help humans understand the black-box, may allow for a deeper human-machine interaction \cite{koh2020concept}, and may lead to more trustworthy fully automated tasks. All of this is possible as long as the explanations are useful from a human perspective.
In a nutshell, an explanation is an answer to a ``why'' question 
and what makes an explanation good or bad depends on ``the degree to which a human can understand the cause of a decision'' \cite{miller2019explanation}. The goodness of an explanation is intimately connected to how humans collect evidences and eventually make decisions.
In Herbert Simon's words we may say that a good explanation is \textit{satisficing} when ``it either gives an optimal description of a simplified version of the black-box (e.g. a surrogate model) or a satisfactory description for the black-box itself''
\cite{simon1979rational}. However, the notion itself of explanation generally depends on both the application domain and whom it is aimed at \cite{carvalho2019machine}.

The need for human-understandable explanations is one of the main reasons why concept-based models are receiving ever-growing consideration, as they provide explanations in terms of human-understandable symbols (the \textit{concepts}) rather than raw features such as pixels or characters \cite{kim2018tcav,ghorbani2019towards,koh2020concept}. As a consequence, they seem more suitable to serve many strategic human purposes such as decision making tasks. For instance, a concept-based explanation may describe a high-level category through its attributes as in ``a \textit{human} has \textit{hands} and a \textit{head}''.
While concept ranking is a common feature of concept-based techniques, there are very few approaches formulating hypotheses on how black-boxes combine concepts to arrive to a decision and even less provide synthetic explanations whose validity can be quantitatively assessed \cite{das2020opportunities}.

A possible solution to provide human-understandable explanations is to rely on a formal language that is
very expressive, closely related to reasoning, and somewhat related to natural language expressions, such as First-Order Logic (FOL). A FOL explanation can be considered a special kind of a concept-based explanation, where the description is given in terms of logic predicates, connectives and quantifiers, such as ``$\forall x:\ \textit{is\_human}(x) \rightarrow \textit{has\_hands}(x) \wedge \textit{has\_head}(x)$'', that reads ``being human implies having hands and head''. However, FOL formulas can generally express much more complex relationships among the concepts 
involved in a certain explanation. Compared to other concept-based techniques, logic-based explanations provide many key advantages, that we briefly describe in what follows.
An explanation reported in FOL is a rigorous and unambiguous statement ({clarity}). This formal clarity may serve cognitive-behavioral purposes such as engendering trust, aiding bias identification, or taking actions/decisions. For instance, dropping quantifiers and variables for simplicity, the formula ``\textit{snow} $\wedge$ \textit{tree} $\leftrightarrow$ \textit{wolf}'' may easily outline the presence of a bias in the collection of training data.
Different logic-based explanations can be combined to describe groups of observations or global phenomena ({modularity}). 
For instance, for an image showing only the face of a person, an explanation could be ``(\textit{nose} $\wedge$ \textit{lips}) $\rightarrow$ \textit{human}'', while for another image showing a person from behind a valid explanation could be ``(\textit{feet} $\wedge$ \textit{hair} $\wedge$ \textit{ears})  $\rightarrow$ \textit{human}''. The two local explanations can be combined into ``(\textit{nose} $\wedge$ \textit{lips}) $\vee$ (\textit{feet} $\wedge$ \textit{hair} $\wedge$ \textit{ears}) $\rightarrow$ \textit{human}''.
The quality of logic-based explanations can be quantitatively measured
to check their correctness and completeness ({measurability}). For instance, once the explanation ``(\textit{nose} $\wedge$ \textit{lips}) $\vee$ (\textit{feet} $\wedge$ \textit{hair} $\wedge$ \textit{ears})'' is extracted for the class \textit{human}, this logic formula can be applied on a test set to check its generality in terms of quantitative metrics like accuracy, fidelity and consistency.
Further, FOL-based explanations can be rewritten in
    different equivalent forms such as in \emph{Disjunctive Normal Form} (DNF) and \emph{Conjunctive Normal Form} (CNF)  ({versatility}). Finally, techniques such as the Quine–McCluskey algorithm can be used to compact and simplify logic explanations \cite{mccoll1878calculus,quine1952problem,mccluskey1956minimization} ({simplifiability}). As a toy example, consider the explanation ``(\textit{person} $\wedge$ \textit{nose}) $\vee$ ($\neg$\textit{person} $\wedge$ \textit{nose})'', that can be easily simplified in ``\textit{nose}''.

This paper presents a unified framework for XAI allowing the design of a family of neural models, the \emph{Logic Explained Networks} (LENs), which are trained to \textit{solve-and-explain} a categorical learning problem integrating elements from deep learning and logic.
Differently from vanilla neural architectures, LENs can be directly interpreted by means of a set of FOL formulas. In order to implement such a property, LENs require their inputs to represent the activation scores of human-understandable concepts. Then, specifically designed learning objectives allow LENs to make predictions in a way that is well suited for providing FOL-based explanations that involve the input concepts. In order to reach this goal, LENs leverage parsimony criteria aimed at keeping their structure simple.
There are several different computational pipelines in which a LEN can be configured, depending on the properties  of the considered problem and on other potential experimental constraints. For example, LENs can be used to directly classify data in an explainable manner, or to explain another black-box neural classifier. Moreover, according to the user expectations, different kinds of logic rules may be provided.
Due to this intrinsic versatility of LENs, what we propose can also be thought as a generic framework that encompasses a large variety of use cases.

\begin{figure}[!t]
    \centering
    \includegraphics[width=1\textwidth]{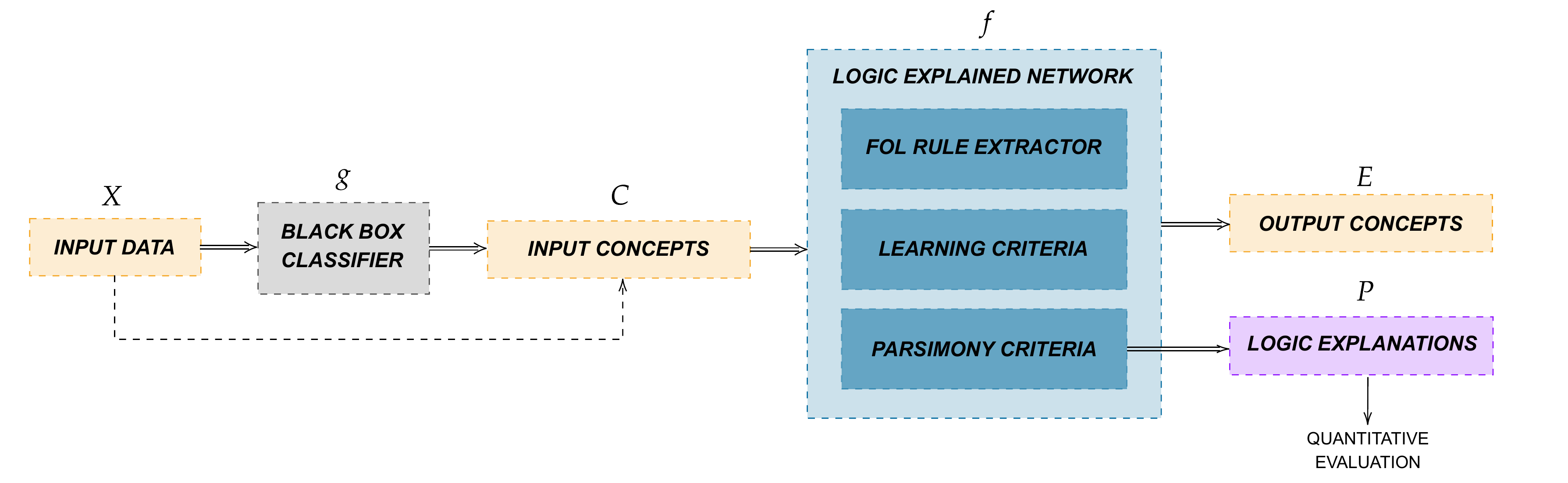}
\vskip -2mm
\caption{Logic Explained Networks (LENs, $f$) are neural networks capable of making predictions of a set of output concepts (activation scores belonging to $E$) and providing First-Order Logic explanations (belonging to $P$) in function of the LEN inputs. Inputs might be other concepts (activation scores belonging to $C$) either computed by a neural network ($g$) or directly  provided within the available data (each data sample belongs to $X$).  
There are several different ways of instantiating this generic model into real-world problems (Section~\ref{sec:basics}).
Within the blue box, the key components of LENs are listed (Section~\ref{sec:methods}).}
    \label{fig:framework}
\end{figure}

\begin{figure}[t]
    \centering
     \includegraphics[width=1.\textwidth, trim = 0 0 35 0, clip]{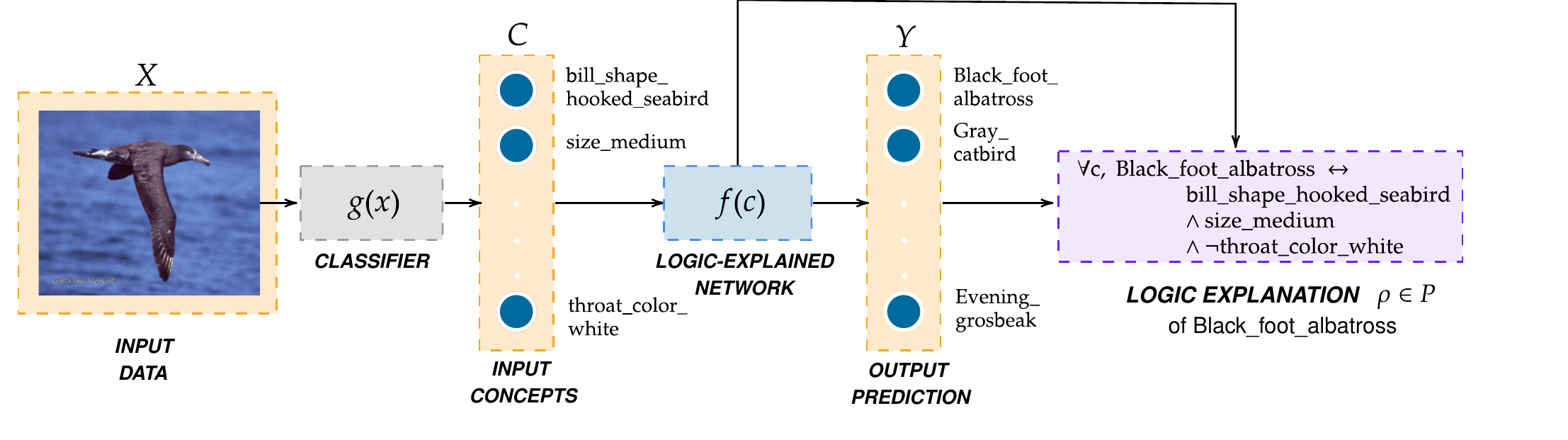}
         \caption{An example of a possible instance of the generic model of Fig.~\ref{fig:framework}, inspired by the CUB 200-2011 fine-grained classification dataset. Classes are divided into a two-level hierarchy. A LEN is placed on top of a convolutional neural network $g(\cdot)$ in order to \textit{(i)} classify the species of the bird in input and \textit{(ii)} provide an explanation on why it belongs to this class. \sm{The logic explanation in the example showcases the predicted output class (all the output concepts can be explained), dropping the argument of the predicates for compactness.}}
         \label{fig:len}
\end{figure}

Fig.~\ref{fig:framework} depicts a generic view on LENs, in which the main components are reported. 
A LEN (blue box -- function $f$) provides FOL explanations (purple box) of a set of output concepts (rightmost yellow-box) in function of the LEN inputs. Inputs might be other concepts (mid yellow box) computed by a neural network classifier (gray box -- function $g$) and/or concepts provided within the available data (leftmost yellow box). 
This generic structure can be instantiated in multiple ways, depending on the final goal of the user and on the properties of the considered problem. In order to provide the reader with an initial example/use-case (different configurations are explored in the paper), we consider an image classification problem with concepts organized into a two level hierarchy. In Fig.~\ref{fig:len} we report an instance of Fig.~\ref{fig:framework} in which a CNN-based neural classifier gets an input image, predicting the activations of a number of low-level concepts.
The LEN $f$ processes such concepts, and it predicts the activation of higher-level output concepts. The LEN can provide a FOL description of each (high-level) output concept with respect to the (low-level) input ones. Another possible instance of the proposed framework consists in using the LEN to directly classify and explain input data (that is basically the case in which the input concepts are immediately available in the data themselves, and not the output of another neural model), thus the LEN itself becomes an interpretable machine. Moreover, LENs can be paired with a black-box classifier operating on the same input data, and forced to mimic as much as possible the behaviour of the black-box, implementing an additional explanation-oriented module. 

We investigate three different use-cases that are inspired by the aforementioned instances, comparing different ways of implementing the LEN models. 
While most of the emphasis of this paper is on supervised classification, we also show how LEN can be leveraged in fully unsupervised settings.
Additional human priors could be eventually incorporated into the learning process \cite{ciravegna2020constraint}, in the architecture \cite{koh2020concept}, and, following Ciravegna et al. \cite{ciravegna2020constraint,ciravegna2020human}, what we propose can be trivially extended to semi-supervised learning (out of the scope of this paper).
Our work contributes to the XAI research field in the following ways.
\begin{itemize}
    \item It generalizes existing neural methods for solving and explaining categorical learning problems \cite{ciravegna2020human,ciravegna2020constraint} into a broad family of neural networks i.e., the \emph{Logic Explained Networks} (LENs).
    \item It describes how users may interconnect LENs in the classification task under investigation, and how to express a set of preferences to get one or more customized explanations. 
    \item It shows how to get a wide range of logic-based explanations, and how logic formulas can be restricted in their scope, working at different levels of granularity (explaining a single sample, a subset of the available data, etc.).
    \item It reports experimental results using three out-of-the-box preset LENs showing how they may generalize better in terms of model accuracy than established white-box models such as decision trees on complex Boolean tasks (in line with Tavares' work \cite{tavares2020understanding}).
    \item It advertises our public implementation of LENs through a Python package\footnote{\url{https://pypi.org/project/torch-explain/}} with an extensive documentation about LENs models, implementing different trade-offs between intepretability/explainability and accuracy.
\end{itemize}

\begin{figure}[!b]
    \centering
    \includegraphics[width=0.98\textwidth]{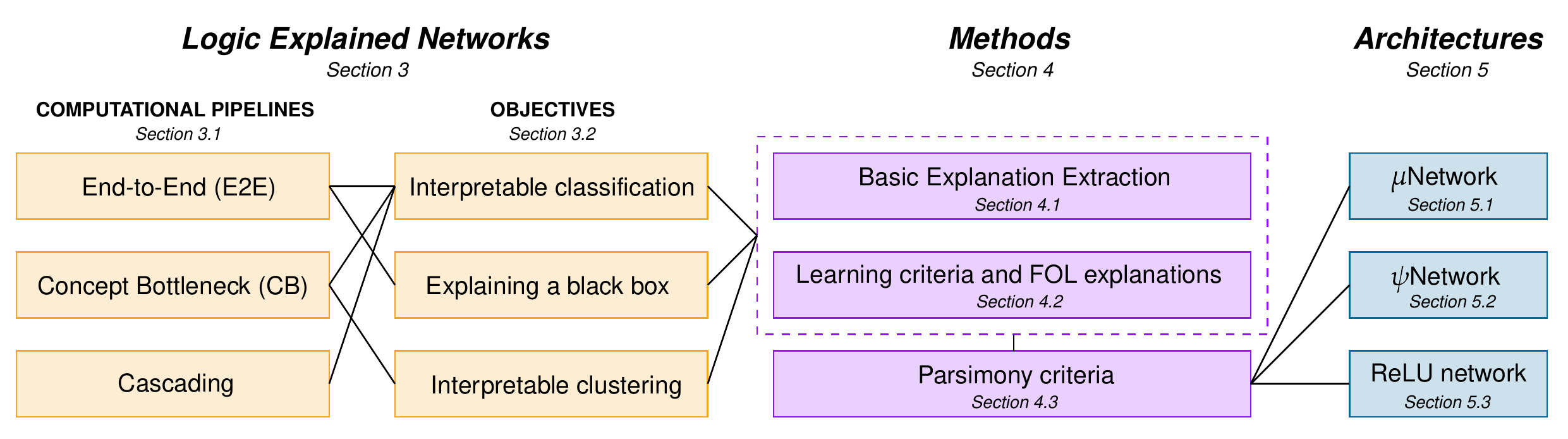}
    \caption{Visual overview of the organization of the paper for those sections that are about describing the whole LEN framework and the specific use-cases we selected. Starting from the nodes on the left, each path that ends to one of the nodes on the right creates a specific instance of the LEN framework. The proposed framework is generic enough to create several other instances than the ones we study in this paper.} 
    \label{fig:visual}
\end{figure}

The paper is organized as follows (see also Fig.~\ref{fig:visual}). Related works are described in Section \ref{sec:related}. Section \ref{sec:basics} gives a formal definition of a LEN and it describes its underlying assumptions, key paradigms and design principles. The methods used to extract logic formulas and to effectively train LENs are described in Section \ref{sec:methods}. Three out-of-the-box LENs, presented in Section \ref{sec:outofthebox}, are compared on a wide range of benchmarks in terms of classification performance and quality of the explanations, in Section \ref{sec:experiments}, \sm{including an evaluation of the LEN rules in an adversarial setting}.
Finally, Section \ref{sec:conclusion} outlines the social and economical impact of this work as well as future research directions.

\section{Related work}
\label{sec:related}

In the last few years, the demand for human-comprehensible models has significantly increased in safety-critical and data-sensible contexts. This popularity is justified by the emerging need for unveiling the decision process of pitch-black models like deep neural networks. To this aim, the scientific community has developed a variety of XAI techniques, with different properties and goals. Several taxonomies have been proposed to categorize the XAI models, with partial overlapping and some ambiguities on the referred terminology. Without pretending to be exhaustive, in the following we focus on some key properties that are relevant to compare LENs with other existing approaches. In particular, we will describe related XAI methods considering the type of \paridelist{result} they provide (feature-scoring vs. rule-based),
their specific \paridelist{role} (interpretable models vs. explanation methods),
and the \paridelist{scope} of the provided explanations (local vs. global). A brief summary of a few XAI works in terms of these features is reported in Tab.~\ref{tab:related_work} -- for more details on existing taxonomies we refer to the recent surveys \cite{adadi2018peeking,carvalho2019machine,marcinkevivcs2020interpretability,molnar2020interpretable}.

XAI models can be differentiated according to the \paridelist{result} of the produced explanation.
Most of the methods in literature usually focus on scoring or providing summary statistics of features  \cite{erhan2010understanding, simonyan2013deep, zeiler2014visualizing, ribeiro2016model, ribeiro2016should, lundberg2017unified, selvaraju2017grad}. However, \textit{feature-scoring} techniques
can be of scarce utility in decision support cases, whereas a comprehensible language may bring light on the black-box behaviour by identifying concept-based correlations.
On the other hand, \emph{rule-based} methods are generally more comprehensible \cite{breiman1984classification, angelino2018learning, letham2015interpretable}, since they usually rely on a formal language, such as FOL. Further, the learned rules can be directly applied to perform the learning task in place of the explained model.
%
Existing approaches have different \textsc{roles} in the XAI landscape, acting as intrinsically \emph{interpretable models} or as \emph{explanation methods}. 
As a matter of fact, the interpretability of a model can be achieved either by constraining the model to be interpretable per se (\emph{interpretable models}) or by applying some methods to get an explanation of an existing model (\emph{explanation methods}). Interpretable models are by definition {model specific}, while explanation methods can rely on a single model or be, as more commonly happens, {model agnostic}. In principle, interpretable models are the best suited to be employed as decision support systems. However, their decision function often is not smooth which makes them quite sensible to data distribution shifts and impairs their generalization ability on unseen observations \cite{tavares2020understanding,molnar2020interpretable}.
On the other hand, explanation methods can be applied to get approximated interpretations of state-of-the-art models. These methods are known as {``post hoc''} techniques, as explanations are produced once the training procedure is concluded.
%
Finally, a fundamental feature to distinguish among explainable methods is the \textsc{scope} of the provided explanations. \textit{Local} explanations are valid for a single sample, while \textit{global} explanations {hold} on the whole input space. However, several models consider local explanations together with aggregating heuristics to get a global explanation.
Some of the most famous XAI techniques share the characteristics of being feature-scoring, local, post hoc methods.


\begin{table}[t]
\centering
\setlength{\tabcolsep}{8pt} 
\renewcommand{\arraystretch}{1.3} 
\resizebox{\textwidth}{!}{
\begin{tabular}{l|cc|cc|cc}
                                    & \multicolumn{2}{c|}{\textbf{Result Type}} & \multicolumn{2}{c|}{\textbf{Scope}} & \multicolumn{2}{c}{\textbf{Role}} \\
                        & feature-scoring & rule-based & local & global & interpret. model & expl. method \\
\hline
LIME &\faCheckCircle & & \faCheckCircle            & & & \faCheckCircle  \\
SHAP &\faCheckCircle & & \faCheckCircle & & & \faCheckCircle  \\
Activation Maximization & \faCheckCircle & & \faCheckCircle  &  &  & \faCheckCircle  \\
Saliency Maps &\faCheckCircle&& \faCheckCircle &      &              & \faCheckCircle \\
SP-LIME &\faCheckCircle&& & \faCheckCircle &              & \faCheckCircle  \\
Class Model Visualization &\faCheckCircle&&              & \faCheckCircle          &              & \faCheckCircle                 \\
LORE &&\faCheckCircle& \faCheckCircle & &  & \faCheckCircle \\
Anchors &&\faCheckCircle& \faCheckCircle & &  & \faCheckCircle  \\
DeepRED   &&\faCheckCircle& \faCheckCircle & \faCheckCircle &  & \faCheckCircle \\
GAM  &\faCheckCircle&&              & \faCheckCircle          & \faCheckCircle            &   \\
Decision Trees       &&\faCheckCircle& \faCheckCircle & \faCheckCircle          & \faCheckCircle            &               \\
BRL                   &&\faCheckCircle& \faCheckCircle & \faCheckCircle          & \faCheckCircle            &               \\
\hline
\textbf{LENs}                                           &&\faCheckCircle& \textbf{\faCheckCircle}    & \textbf{\faCheckCircle}  & \textbf{\faCheckCircle}  & \textbf{\faCheckCircle} \\
\end{tabular}
}
\vspace{0.3cm}
\caption{Summary of related work. The first column lists a number of approaches in the context of XAI. The other columns are about different properties. See the paper text for more details. }
\label{tab:related_work}
\end{table}

Sometimes the easiest way to solve the explanation problem is simply to treat the model as a black-box and, sample-by-sample, determine which are the most important features for a prediction.
Prominent examples of algorithms falling in this area 
accomplish this task by perturbing the input data.
Local Interpretable Model-agnostic Explanations (LIME) \cite{ribeiro2016should} trains a white-box model (e.g. a logistic regression) to mimic the predictions of a given model in the neighbourhood of the sample to explain. By analyzing the weights of the white-box model it identifies the most important group of pixels (superpixel).
Differently, SHapley Additive exPlanations (SHAP) \cite{lundberg2017unified} computes the Shapley value of each feature by removing each of them in an iterative manner. 
Other important techniques 
provide the same type of explanations through gradient analysis. 
For instance, Erhan, Courville, and Bengio introduced the Activation Maximization \cite{erhan2010understanding} framing this as an optimization problem. They explain the behaviour of a hidden/output unit by slightly modifying a given input data such that it maximizes its activation. An easier way for identifying the most relevant input features consists in computing Saliency Maps \cite{simonyan2013deep}, which backtrack the classification loss back to the input image.
%
Other post hoc methods have been devised to extract global feature-scoring explanations, often starting from local explanations. 
A submodular-pick algorithm extends the LIME approach (SP-LIME) 
to provide global explanations \cite{ribeiro2016should}. SP-LIME first finds superpixels of all input samples with the standard LIME procedure. Successively, it identifies the minimum number of common superpixels covering most of the images.
Starting from the idea of Activation Maximization, Class Model Visualization \cite{simonyan2013deep} searches the input sample maximizing class probability scores in the whole input space. 
While inheriting several of the above properties, LENs take a different perspective, as they aim at providing human-comprehensible explanations in terms of FOL formulas (see Tab.~\ref{tab:related_work}).

Existing XAI methods can also supply FOL explanations.
LOcal Rule-based Explanations of black-box decision systems (LORE) \cite{guidotti2018local} extracts FOL explanations via tree induction. Here the authors focus on local rules extracted through input perturbation. For each sample, a simple decision tree is trained to mimic the behaviour of the black-box model in the neighbourhood of the sample. Both standard and counterfactual explanations can then be extracted  following the branches of the tree.
Also the Anchors method \cite{ribeiro2018anchors}, based on LIME, provides local rules of black-box model predictions via input perturbation. In this case, rules are not extracted from a decision tree but are generated by solving a Multi-Armed-Bandit beam search problem.
Extending the CRED algorithm \cite{sato2001rule}, DeepRED  \cite{zilke2016deepred} employs a decomposition strategy to globally explain neural networks. Starting from the output nodes, the predictions of each neuron are explained in terms of the activations of the neurons in the previous layer by training a decision tree. In the end, all rules for each class are merged in a single global formula in terms of input features. For a given sample, a unique local rule is extracted following the firing path.
The approach proposed in this paper resembles the DeepRED algorithm. As it will become clear in the following sections, LENs can explain the behaviour of a neural network both end-to-end and layer by layer. However, LENs can be used to explain any black-box model as far as its input and output correspond to human-interpretable categories (Section~\ref{sec:basics}). Furthermore, LENs support different forms of FOL rules, with different scopes and goals.

Interpretable models are capable of providing both local and global explanations. Such explanations may be based either on feature rankings, as in Generalized Additive Models (GAM) \cite{hastie1987generalized}, or on logic formulas, as in Decision Trees \cite{breiman1984classification} or Bayesian Rule Lists (BRL) \cite{letham2015interpretable}. GAMs overcome the linearity assumption of linear regression by learning target categories disjointly from each feature. Caruna et al. proposed GA$^2$M where pairs of features are allowed to interact \cite{caruana2015intelligible}. A further extension employing neural networks as non-linear functions has recently been proposed by Agarwal et al. \cite{agarwal2020neural}, which, however, loses interpretability at feature-level.
Decision trees \cite{breiman1984classification} are a greedy learning algorithm partitioning input data into smaller subsets until each partition only contains elements belonging to the same class. Different pruning algorithms have been proposed to simplify the final structure to get simple explanations \cite{quinlan1987simplifying}. Each path of a decision tree is equivalent to a decision rule, i.e. an \textit{IF-THEN} statement.
Other approaches focus on generating sets of decision rules, either with sequential covering \cite{cohen1995fast} or by selecting the best rules from a pre-mined set via Bayesian statistics \cite{letham2015interpretable}.
Interestingly, LENs can be used to solve a learning task directly, as they are interpretable per se.
Employing an interpretable-by-design neural network has the advantage that the extracted explanations will perfectly match the classifier predictions.
In addition, relying on a neural network, LEN generalization performance can be much better than standard interpretable-by-design methods, whose representation capacity is usually limited.

To sum up, the proposed family of neural networks can be used both as interpretable classifiers and as surrogate models. Indeed, the flexibility of the proposed framework allows the user to find an appropriate trade-off between interpretability/explainability and accuracy that is well suited for the task at hand.
Concerning the explanation type, LENs provide explanations as FOL formulas. The inner mechanism to extract such explanations is general enough to cover rules with different scopes, from local rules valid on a single sample to global rules that hold for an entire class.

\section{Logic Explained Networks}
\label{sec:basics}


This work presents a special family of neural networks, referred to as Logic Explained Networks (LENs), which are able to both make predictions and provide explanations. LENs may explain either their own predictions or the behaviour of another classifier, being it a neural network or a generic black-box model. Moreover, a LEN might also be used to explain relationships among some given data.
In this section, we describe how LENs can be instantiated in the framework of Fig.~\ref{fig:framework}. We start by introducing the notation (following paragraphs). Then we discuss a variety of computational pipelines (Section~\ref{sec:use_cases}) and illustrate different types of explanations LENs can generate in function of different learning objectives (Section~\ref{sec:explanation_styles}). 
In Fig.~\ref{fig:visual} (leftmost part) we provide an overview of the contents of this section. In the same figure, we also show how the following sections will cover the specific methods (Section~\ref{sec:methods}) and the considered neural architectures (Section~\ref{sec:outofthebox}).

A LEN is a function $f$, implemented with a neural network, that maps data falling within a $k$-dimensional Boolean hypercube onto data falling within another $r$-dimensional Boolean hypercube. Each dimension is about the activation strength of what we refer to as a \textit{concept}, with the strict requirement of having a human-understandable description of each \textit{input} dimension/concept \cite{kim2018tcav}.
Formally, a LEN is a mapping $f: C \rightarrow E$, where $C = [0,1]^{k}$ is the \textit{input concept space}, and $E = [0,1]^{r}$ is the so called 
\textit{output concept space}. 
\sm{FOL explanations produced by a LEN are about the relationships between the output and the input concepts, and they belong to the generic \textit{rule space} $P$.
In particular, whenever a LEN has been trained, the $i$-th output 
$f_i$
can be directly translated into a logic rule $\varphi_i$ that involves the input concepts and that is leveraged to devise a FOL formula $\rho_i \in P$, such as the one shown in Fig.~\ref{fig:len}. The FOL extraction process is general enough to offer logic rules with different levels of granularity, from a local to a more global coverage of the available data.}

In order to provide a concrete meaning to LEN models, the source of the input concept activations needs to be defined as well as the learning criteria. Amongst a variety of possible configurations that are instance of Fig.~\ref{fig:framework}, we will focus on a limited set of computational pipelines in which the input and output concept spaces are defined with respect to real-world use cases. In some of them, LENs communicate with a black-box classifier, with the goal of providing explanations of its predictions, or with the goal of leveraging its activations as input concepts.
In all cases, for each input sample, LENs provide $r$ FOL explanations that might either be directly associated to $r$ categories of a classification problem or they could be interpreted as $r$ explanations of unknown relationships among input data. In the former case, LENs are trained in a supervised manner, while in the latter case they are trained in an unsupervised setting.

Before going into further details, we introduce the main entities and the notation that will be used throughout the paper, \sm{paired with a short description to help the reader in following the paper and to have a quick reference to the main symbols.}
\begin{itemize}
\item[$f$, $C$, $E$:] The function computed by a LEN is $f \colon C \rightarrow E$, where $C = [0,1]^{k}$ is the space of the activations of the $k$ input concepts, and $E = [0,1]^{r}$ is about the activations of the $r$ output concepts.
\item[$X$, $\mathcal{X}$, $\mathcal{C}$:] We consider a scenario in which a finite collection $\mathcal{X}$ of data samples that belong to $X\subseteq\mathbb{R}^d$ is available to train LENs. \sm{There exists a mapping from $X$ to the space $C$ of input concept activations. The notation $\mathcal{C}$ indicates the finite set of concept activations obtained by mapping each sample of $\mathcal{X}$ onto $C$.}
    \item[$q$:] We use the notation $q$ to indicate the total number of classes in a classification problem. Each data sample can be associated with one or more classes. No strict conditions are posed on the potential mutual exclusivity of the different classes, thus we consider the most generic setting that spans from multi-label to single-label classification. Moreover, classes could also be organized in a structured manner, such as in hierarchy. 
    \item[$\bar{y}$, $\bar{y}_i$:] \sm{Whenever we consider classification problems, classes are assumed to be} encoded with binary targets in $\{0,1\}^{q}$. The function $\bar{y}(\cdot)$ returns the target vector associated to the data sample passed as its argument, while $\bar{y}_i(\cdot)$ returns the \sm{$i$-th component of such a vector, that is about the $i$-th class.} \sm{What we propose holds also in the case in which targets are encoded with scores in the unit interval, so that we will frequently refer to generic data encoded in $Y=[0,1]^{q}$.} 
 \item[$Y^{(a:b)}$:] \sm{We will use the notation $Y^{(a:b)}$ to indicate a view of $Y$ limited to the dimensions going from index $a \geq 1$ to index $b \leq q$, included. }
\item[$g$:] The function $g(x)$ is about a generic black-box neural classifier \sm{that computes the membership scores of $x$ to a set of categories. Such categories could be exactly the ones of the considered classification problem (encoded in $Y$) or a subset of them (encoded in $Y^{(a:b)}$).}
 In detail, $g: X \rightarrow Y$ \sm{(resp. $g: X \rightarrow Y^{(a:b)}$)}, and $g(x)$  defines the membership scores of $x$ to the considered categories.
 Of course,  $g_i(x)=1$ when $x$ strongly belongs to class $i$ \sm{(resp. $a+i-1$)}. No special conditions are enforced in the definition of $g$. 
 \item[$\bar{f}$, $\bar{f}_i$:] The notation $\bar{f}$ is used to indicate a Boolean instance of \sm{the main LEN} function $f$, in which each output of $f$ is projected to either $0$ or $1$.\footnote{When not directly specified, we will assume $0.5$ to be used as threshold value to compute the projection.} In order to refer to a single output of the vector function $f$ (or $\bar{f}$), the subscript will be used, e.g. $f_i$ (or $\bar{f}_i$). 
 \item[$c_j$, $\bar{c}_j$:] Similarly, for a vector $c \in C$ with activation scores of $k$ concepts, $c_j$ is the $j$-th score, while $\bar{c}_j$ is the Boolean instance of it. 
 \item[$\bar{c}_j$ {\scriptsize (name)}:] Any dimension of the space of input concepts $C$ includes a human-readable label. Logic rules generated by LENs will leverage such labels to give understandable names to predicates. In order to simplify the notation, whenever we \sm{report} a logic rule, we will use the already introduced symbol $\bar{c}_j$ also to refer to the human-understandable name of \sm{the $j$-th input concept. This notation clash makes the presentation easier.}
 \item[$\bar{y}_i$ {\scriptsize (name)}:] \sm{Any dimension of the space of output concepts $E$ might or might not include a human-readable label, whether we consider supervised or unsupervised learning when training LENs, respectively. In the former case, the already introduced notation $\bar{y}_i$ will also refer to the human-understandable name of the $i$-th output class, following the same simplification we described in the case of $\bar{c}_j$.}
 \item[$\varphi_i$:] \sm{Any output $f_i$ is associated with a logic rule $\varphi_i$ \sm{given in terms of (the names of) the input concepts}.}
 \item[$\rho_i$:] \sm{We indicate with $\rho_i \in P$ the FOL formula that explains the $i$-th output concept leveraging $\varphi_i$. The precise way in which $\varphi_i$ is used to build $\rho_i$ depends on whether we are considering supervised or unsupervised output concepts.}
\end{itemize}
The listed elements will play a precise role in different portions of the LEN framework.

\begin{figure}[t]
    \centering
    $\vcenter{\hbox{\includegraphics[trim=0 0 0 0,clip,width=.6\textwidth]{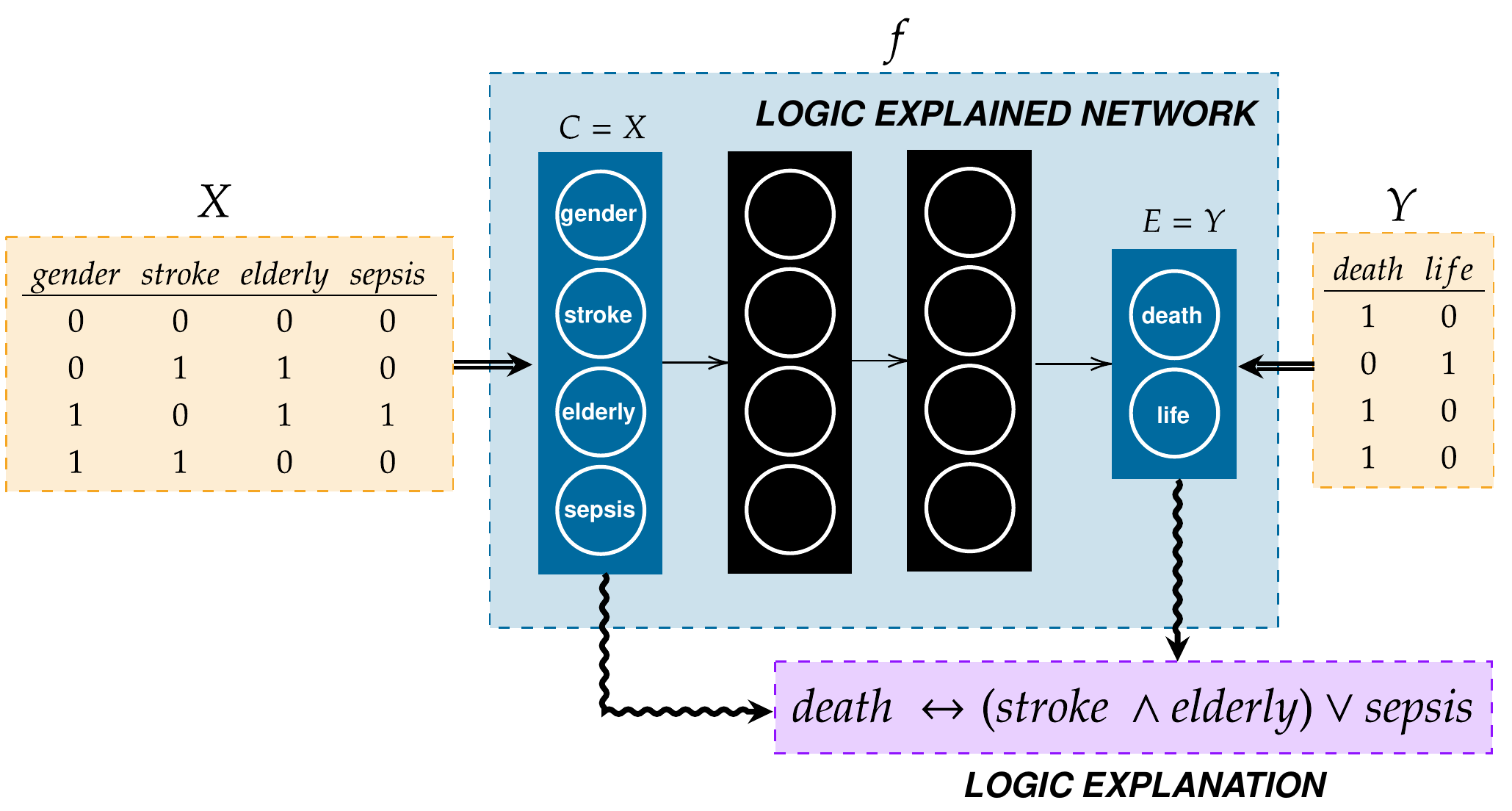}}}$ \\ 
    $\vcenter{\hbox{\includegraphics[trim=0 0 0 0,clip,width=.7\textwidth]{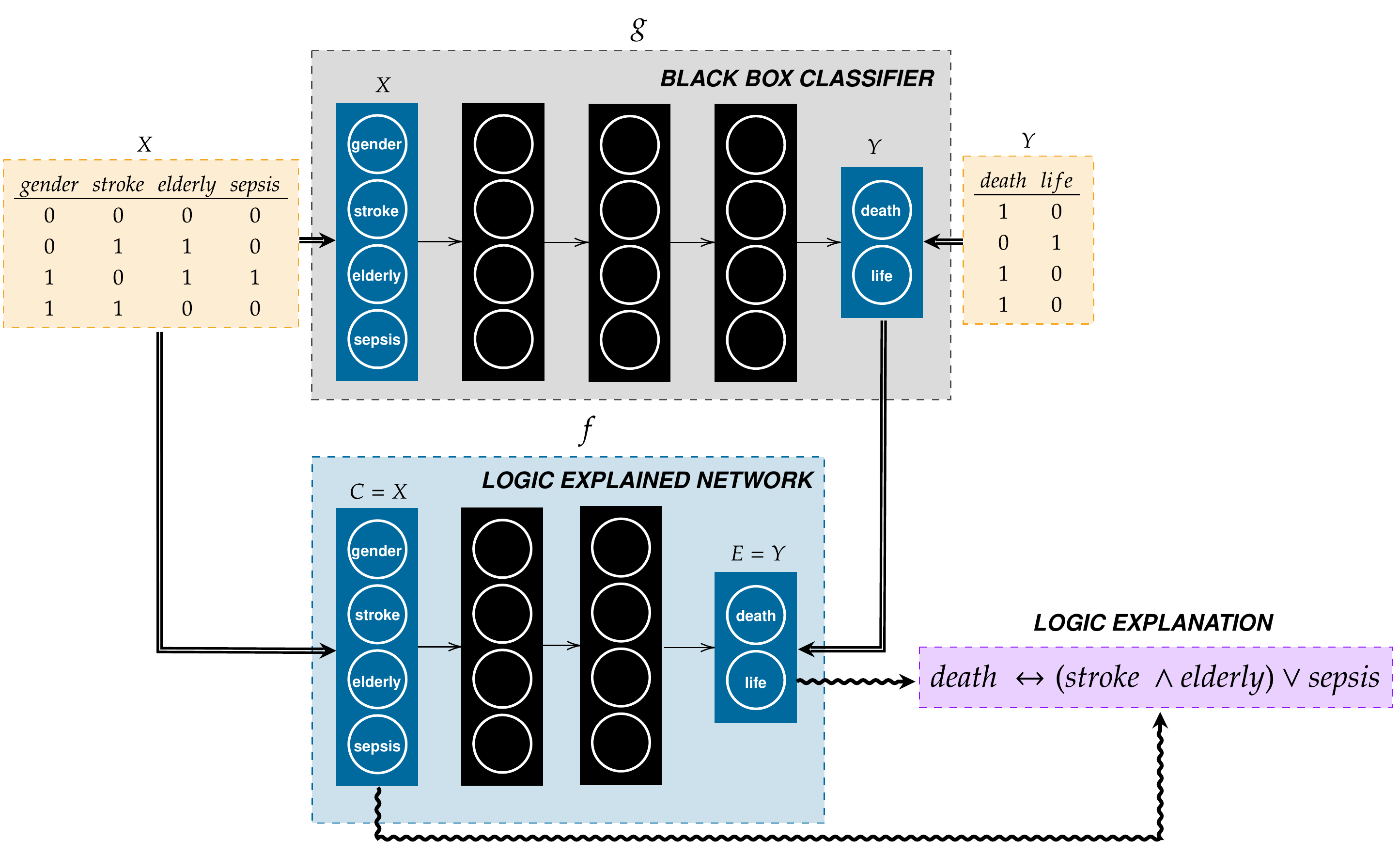}}}$
    \caption{End-to-end (E2E) LENs directly work on input data that are interpretable per se and treated as concepts ($C=X$), while the output concepts are the activation scores of the classes of the dataset ($r = q = 2$). This is a real-world example from the MIMIC II dataset (see the experimental section), where patient features are used to classify if the patient will survive 28 days after hospital recovery. Top: the LEN solves the classification problem and provides explanation, also referred to as \textit{interpretable classification}. Bottom: the LEN \textit{provides explanations} of a black-box classifier. The universal quantifier and the argument of the predicates have been dropped for simplicity.}
    \label{fig:E2EExplainer}
\end{figure}

\subsection{Computational pipelines}
\label{sec:use_cases}

Amongst a large set of feasible input/output configurations of the LEN block, here we consider three different computational pipelines fitting three concrete scenarios/use-cases where LENs may be applied, that consist in what we refer to as \paridelist{end-to-end}, \paridelist{concept bottleneck}, and \paridelist{cascading} pipelines.



\paride{End-to-end (E2E).} The most immediate instance of the LEN framework is the one in which the LEN block aims at directly explaining, and eventually solving, a supervised categorical learning problem. In this case, we have $C = X$, and $r = q$, as shown in Fig.~\ref{fig:E2EExplainer}. In order to make the first equality valid with respect to the LEN requirements, the $d$ input features of the data in $X$ must score in $[0,1]$ (or in $\{0,1\}$ in the most extreme case), so that they can be considered as activations of $d$ human-interpretable concepts. 
In order to create these conditions in different types of datasets, generic continuous features can be either discretized into different bins or Booleanized by comparing them to a reference ground truth (e.g. for gene expression data, the gene signature obtained from control groups can be used as a reference).
($i$) An E2E LEN can perform a fully \textit{interpretable classification} task, having the double role of main classifier and explanation provider (Fig.~\ref{fig:E2EExplainer}, top), or ($ii$) it can work in parallel with another neural black-box classifier $g$, thus {providing explanations} of the predictions of $g$ (Fig.~\ref{fig:E2EExplainer}, bottom) -- what we refer to as \textit{explaining a black-box}.
\sm{In both the cases ($i$) and ($ii$) we have} $E = Y$. 
The first solution is preferred when a white-box classifier is of utmost importance, while a loss in terms of classification accuracy is acceptable, due to the constrained nature of the LENs.
The second solution is useful in contexts where the classification performance plays a key role, and approximate explanations of black-box decisions are sufficient.

\begin{figure}[t]
\centering
    $\vcenter{\hbox{\includegraphics[width=.6\textwidth]{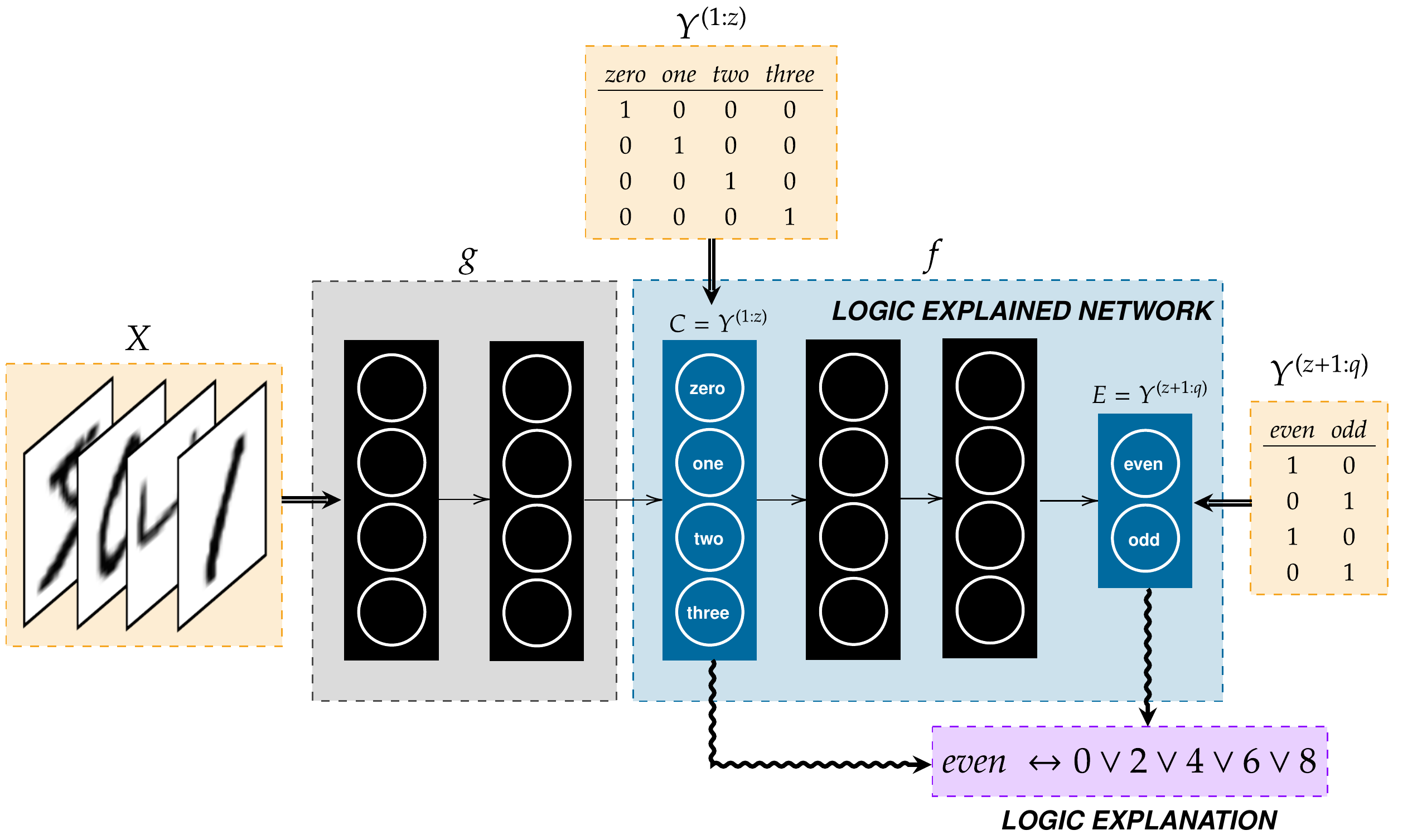}}}$\\ 
    $\vcenter{\hbox{\includegraphics[width=.6\textwidth]{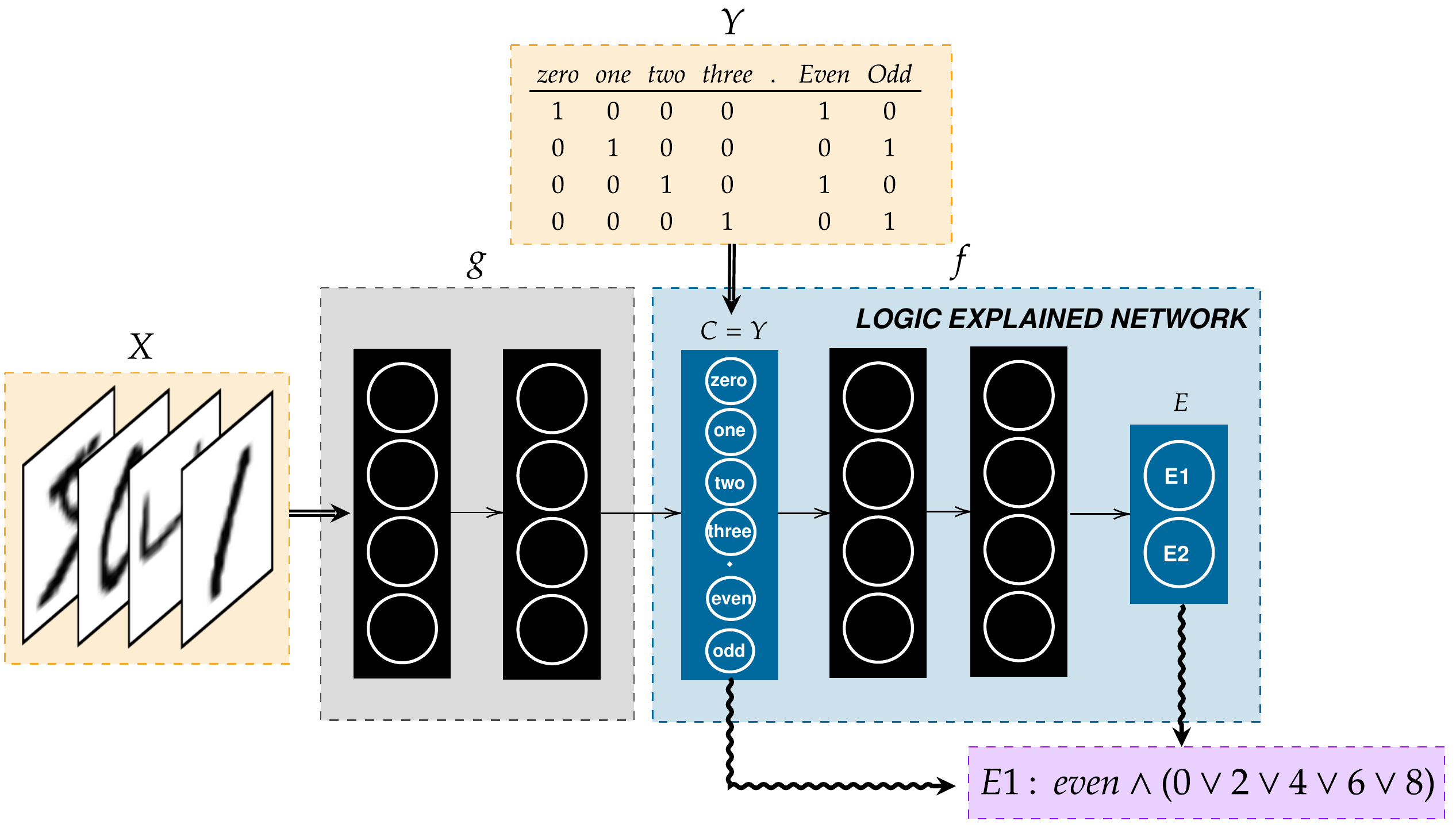}}}$
    \caption{Concept-bottleneck (CB) LEN. Left: the LEN is placed on top of a black-box model $g$ which maps the input data into a first set of interpretable concepts. An MNIST-based experiment is shown (see the experimental section). Handwritten digits are first classified by $g$. A LEN is then employed to classify and explain whether the digit is even or odd -- \textit{interpretable classification}.  Top: MNIST digits are classified as belonging to one of the 10 classes and whether they are even of odd by a black-box $g$ (see the experimental section).  
     Bottom: A LEN groups these predictions in an unsupervised manner, and analyzes the relations within each cluster -- \textit{interpretable clustering}. Supervision labels are not provided for the LEN output. The universal quantifier and the argument of the predicates have been dropped for simplicity. 
     }
    \label{fig:CMBExplainer}
\end{figure}

\paride{Concept-bottleneck (CB).} A Concept-Bottleneck LEN is a computational pipeline in which the LEN $f$ aims at explaining, and eventually solving, a categorical learning problem whose features do not correspond to human-interpretable concepts and, as a consequence, they are not suitable as LEN inputs. In this case, a black-box model $g$ computes the activations of an initial set of $z$ concepts out of the available data $\mathcal{X}$. The LEN solves the problem of predicting $r$ new concepts from the activations of the $z$ ones, as shown in Fig.~\ref{fig:CMBExplainer} (top/bottom). Differently from the E2E case, there is always a neural network $g$ processing the data, and the outcome of such processing is the input of the LEN. Formally, we have $g: X \rightarrow Y^{(1:z)}$, while $f: C \rightarrow E$ with $C = Y^{(1:z)}$, with $z \leq q$,\footnote{\sm{We implicitly assumed the axes of $Y$ to be sorted so that the first $z$ dimensions are the ones we want to predict with $g$, and we will make this assumption in the whole paper.}} and where the meaning of $E$ varies in function of what we describe in the following.
($i$) This two-level scenario can be implemented in a way that is coherent with the already discussed example of Fig.~\ref{fig:len}, that is what we also show in Fig.~\ref{fig:CMBExplainer} (top). In this case, each example is labeled with $q$ binary targets divided into two disjoint sets, where the black-box network $g$ predicts the first $z < q$ ones, and the LEN $f$ predicts the remaining $r = q-z$, i.e. $E = Y^{(z+1:q)}$.
The LEN will be both responsible of predicting the higher level concepts and of explaining them in function of the lower level concepts yielded by the black-box model, thus still falling withing the context of \textit{interpretable classification}.
 ($ii$) We also consider the case in which the outcome of the LEN is not associated to any known categories. In this case, the LEN is trained in an unsupervised manner, as shown in  Fig.~\ref{fig:CMBExplainer} (bottom), thus implementing a form of \textit{interpretable clustering}. \sm{Formally, $g: X \rightarrow Y$, $C = Y$, and $E = [0,1]^{r}$, with customizable $r \geq 1$.}
 In both the cases ($i$) and ($ii$), the black-box classifier $g$ is trained in a supervised manner.


\begin{figure}[t]
    \centering
    \includegraphics[trim=0 20 0 0,clip,width=0.7\textwidth]{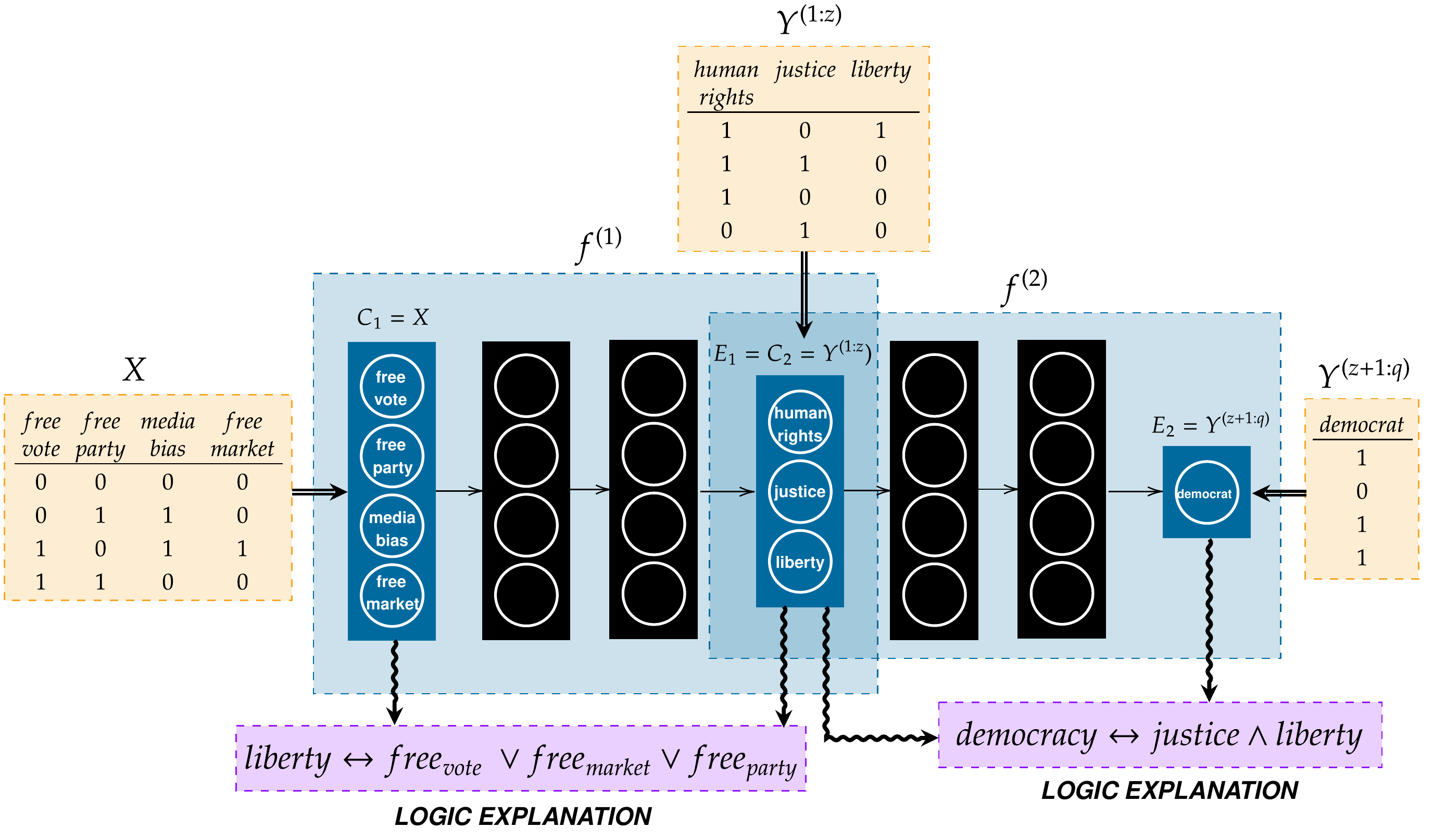}
      \caption{Cascading LENs, an example taken from the V-Dem classification dataset (see the experimental section). The final classification on the status of the democracy is divided into two steps.
    First, the input data/concepts $C_1$ are mapped into high-level concepts $C_2$. 
    High-level concepts are then used to compute the final classification. 
    Cascading LENs can provide explanations at different levels of granularity, implementing multiple \textit{interpretable classifications}. The universal quantifier and the argument of the predicates have been dropped for simplicity.}
    \label{fig:CascadeExplainer}
\end{figure}

\paride{Cascading.} Cascading LENs can be used to provide explanations by means of a hierarchy of concept layers. In particular, multiple LENs are used to map concepts into a higher level of abstraction (without the need of any black-box $g$), 
as shown in Fig.~\ref{fig:CascadeExplainer}. Each LEN provides explanations of its outputs with respect to its inputs, allowing LENs to handle multiple levels of granularity. This structure provides a hierarchy of explanations and upholds human interventions at different levels of abstraction, providing a far deeper support for human-machine interactions. We indicate with $(C_1, E_1), (C_2, E_2), \ldots, (C_s, E_s)$ the input/output concept spaces of the $s$ cascading LENs, with $C_{j} = E_{j-1}$, $j > 1$. In the experiments, we will consider the case in which LENs are trained in a supervised manner, thus implementing multiple \textit{interpretable classifications}.

\subsection{\sm{Objectives}}
\label{sec:explanation_styles}
In Section~\ref{sec:use_cases}, when describing the selected computational pipelines, we made an explicit distinction among \paridelist{interpretable classification}, \paridelist{explaining a black-box}, and \paridelist{interpretable clustering}, as they represent three different \sm{``objectives''} the user might consider in solving the problem at hand. These \sm{objectives} impose precise constraints in the selection of the learning criteria and on the form of the FOL formulas that can be obtained by the LENs. The form of FOL formulas \sm{will be described in detail in Section~\ref{sec:logic_rule_extr} (generic process of logic rule extraction), and in Section~\ref{sec:learncrit} (learning criteria and FOL rules)}.
\sm{The key difference among the aforementioned objectives is about the criterion that drives the learning dynamics of $f$, as each LEN module can be trained in a supervised or an unsupervised fashion.}

\sm{\paride{Interpretable classification.} Whenever LENs are leveraged to both solve the classification problem and provide explanations, i.e. in \paridelist{interpretable classification},  
learning is driven by supervised criteria. Following Ciravegna et al. \cite{ciravegna2020human}, supervised criteria leverage the available data so that: (i) $f$ learns class labels as in classic supervised learning, and (ii) $f$ is constrained to be coherent with the target type of FOL rules. Each output neuron of the LEN is associated to a target category named $\bar{y}_i$ of the considered learning problem. LENs can provide a FOL explanation $\rho_i \in P$ of such category, so that the class-predicate $\bar{y}_i(\cdot)$ will be involved in the extracted FOL formula. For example, if \textit{person} is one of the output categories, LENs can explain the reasons behind the prediction of class $\bar{y}_i =$ \textit{person} by means of an automatically discovered relationship $\varphi_i$ involving the  activations/not-activations of some ($m_i$) of the $k$ input concepts ($m_i < k$). If such input concepts are  $\bar{c}_j=$ \textit{head},  $\bar{c}_z=$ \textit{hands}, $\bar{c}_h=$ \textit{body}, the system could learn that $\rho_i = \forall c \in C\colon \bar{y}_i(c) \leftrightarrow \varphi_i(c)$, where $\varphi_i(c) = \bar{c}_j(c) \land \bar{c}_z(c) \land \bar{c}_h(c)$, i.e. (discarding the quantifier) \textit{person} $\leftrightarrow$ \textit{head} $\land$ \textit{hands} $\land$ \textit{body}. Interestingly, in the case of interpretable classification, LENs basically become 
 white-box classifiers, as is showcased by the concrete examples of Fig.~\ref{fig:E2EExplainer} (top), Fig.~\ref{fig:CMBExplainer} (top), and Fig.~\ref{fig:CascadeExplainer}.}

\paride{Explaining a black-box.} In case the user aims at \paridelist{explaining a black-box}, LENs act in parallel to black-box classifiers with the goal of explaining the decisions of such black-box models. In this scenario, LENs are constrained to mimic the predictions of the black-boxes, i.e. $f$ is forced to be close to $g$ when evaluated on the available data samples. This is what is shown in  Fig.~\ref{fig:E2EExplainer} (bottom) and it may not require any supervision, since learning can be driven by a coherence criterion between the outputs of $g$ and those of $f$ when processing the same data. \sm{The type of FOL rules LENs can extract are the same of the case of interpretable classification, thus the same principles are followed.}

\sm{\paride{Interpretable clustering.} Differently, LENs can be trained using unsupervised criteria, whenever the user is not aiming at explaining the target classes of a classification problem,} but is interested in discovering \sm{generic} relations among input concepts. We refer to this objective as \paridelist{interpretable clustering}.
The user might be interested in discovering  co-occurrences \sm{$\varphi_i$} of input concept activations in not-defined-before subsets $O_i$ of the concept space. \sm{This leads to FOL rules such as $\rho_i = \forall c \in O_i \colon \varphi_i(c)$, where, for example, $\varphi(c) = \bar{c}_e(c) \lor \bar{c}_t(c)$ and $\bar{c}_e = soccer\_ball$, $\bar{c}_t = foot$, and the set $O_i$ can be thought as a cluster}.
\sm{In this scenario, LEN's output neurons are not associated to any known categories (in contrast to previous objectives), but to unspecified generic symbols}, as shown in Fig.~\ref{fig:CMBExplainer} (bottom). The activation score of the output concept represents a cluster membership score and it is \sm{used to define whether an input pattern belongs to the subset $O_i$ or not.}


\sm{In all the discussed objectives, logic formulas are extracted from LENs using the same principles. Rules are then instantiated into FOL formulas that well cope with objective-specific learning criteria. As it will become clear in the following section, due to such generality of the extraction mechanisms}, the user might decide to automatically discover definite regularities focused on single examples, groups of data points, or, more generally, the whole dataset, moving from local to more global explanations.  

\section{Methods}
\label{sec:methods}
This section presents the fundamental methods used to implement Logic Explained Networks \sm{introduced in Section~\ref{sec:basics}}. 
\draft{We start by describing the procedure that is used to extract logic rules out of LENs for an individual observation  or a group of samples (Section~\ref{sec:logic_rule_extr}). This procedure is common to all LENs' \sm{objectives}/use-cases. Then, we provide the formal definition of the learning objectives constraining LENs to provide required types of \sm{FOL} explanations  (Section~\ref{sec:learncrit}).}
Finally, we discuss how to constrain LENs to yield concise logic formulas. To this aim, ad-hoc parsimony criteria (Section~\ref{sec:parsimony}) \sm{are employed in order to bound the complexity of the explanations}. In Fig.~\ref{fig:visual} (middle) we provide an overview of the contents of this section.

\subsection{\draft{Extraction of logic explanations}}
\label{sec:logic_rule_extr}

Once LENs are trained, a logic formula can be associated to each output concept $f_i$. As it will become clear shortly, extracting logic formulas out of trained LENs can be done by inspecting its inputs/outputs accordingly to their Boolean interpretation. 
\sm{We have already introduced the notation $\varphi_i$ to indicate the logic explanation of the output concept $i$. This generic notation will be properly formalized in the following.}
We \sm{overload the symbol $\varphi_i$ to explicitly indicate, when needed, the data subset where the logic explanation holds true, using the notation} $\varphi_{i,\cdot}$. 
Here the second subscript can refer either to a single data sample $c$, $\varphi_{i,c}$, or to a \draft{set} $S$ of data samples, $\varphi_{i,S}$. 
\draft{In practice, $S$ denotes the region of the concept space that is covered by the $i$-th explanation, i.e. the set of concept tuples for which the formula $\varphi_{i,S}$ is true. By aggregating over multiple samples, the scope of the logic formula may be tuned from strictly local \paridelist{example-level explanations} ($S=\{c\}$) to \paridelist{set-level explanations} ($S\subseteq \mathcal{C}$), where the latter can be focused on a precise class, i.e., \paridelist{class-level explanations}. Eventually, for $S=\mathcal{C}$, global logic formulas holding on the whole concept space $\mathcal{C}$ can be extracted.}



\begin{figure}
    \centering
    \includegraphics[width=1\textwidth]{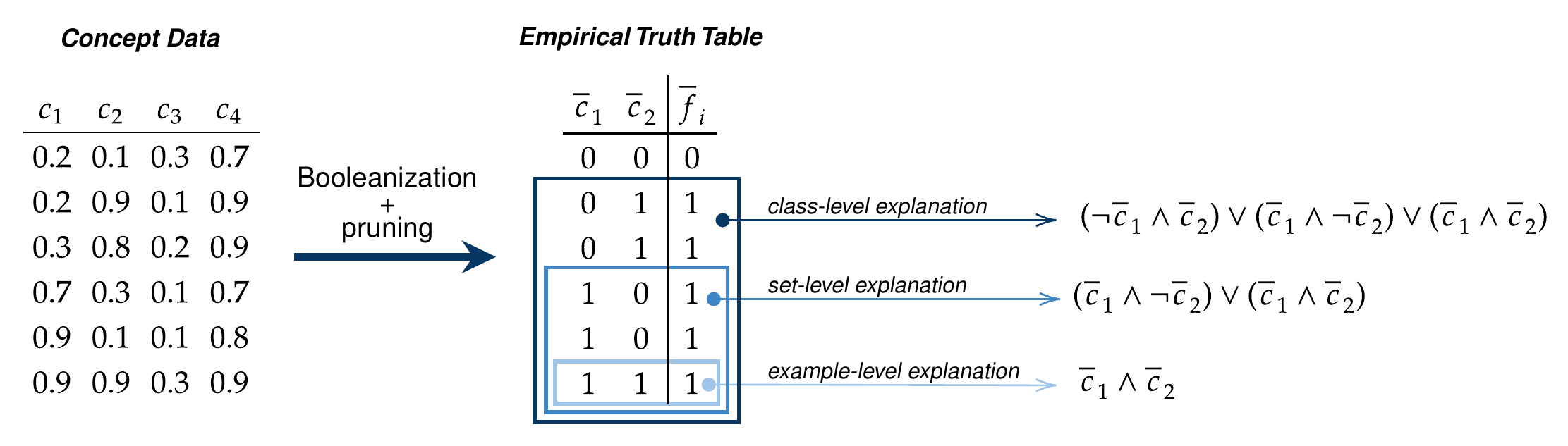}
    \caption{Empirical truth table $\mathcal{T}^i$ of the $i$-th LEN output $f_i$, with $k=4$ and $m_i=2$ (assuming that only the first two input concepts are kept). The aggregation of concept tuples with the same Booleanization yields a common example-level explanation, and do not complicate the class-level explanation. For any example-level explanation, we may count how many samples it explains, to discard  the most infrequent cases.
    }
    \label{fig:tt2form_freq}
\end{figure}

To allow the extraction of FOL formulas, any LEN $f=(f_1,\ldots,f_r)$ requires both its inputs and outputs to belong to the real-unit interval. 
This apparent limitation allows any $f_i$, for $i=1,\ldots,r$, to correspond to \draft{a logic formula.
First, \sm{$f$ maps the data in $\mathcal{C}$ into the rule space $E$. After this forward pass, both the input data $\mathcal{C}$ and the predictions of $f$ are thresholded}, e.g. with respect to $0.5$, to obtain their Boolean values. Then, \sm{for each output neuron $i$}, an \textit{empirical truth-table} $\mathcal{T}^i$ is built by concatenating the $k$-columns of Booleanized input concept tuples $\{ \bar{c},\ \forall c\in\mathcal{C}\}$, with the column of the corresponding LEN's predictions $\bar{f}_i(c)$ (left-side of Fig. \ref{fig:tt2form_freq}). 
The truth-table $\mathcal{T}^i$ can be converted into a logic formula $\varphi_i$ in Disjunctive Normal Form (DNF) as commonly done in \sm{related} literature \cite{mendelson2009introduction}.
However, \sm{the rationale behind LENs is to extract formulas that are compact, emphasizing the most relevant relationships among the input concepts, according to specific parsimony indexes (that will be the subject of Section \ref{sec:parsimony}). Thus,} any $f_i$ will depend only on a proper subset of $m_i \leq k$ concepts, and the formula $\varphi_i$ will be built according to the restriction of $\mathcal{T}^i$ to \sm{$m_i \leq k$} columns (see e.g. Fig.~\ref{fig:tt2form_freq}). Notice that, for convenience in the notation, we assumed the first $m_i$ columns to be the ones playing a role in the explanation, even if they could be any set of $m_i$ columns of $\mathcal{T}^i$.}
In order to give more details about the rule extraction, we \sm{formally} introduce the set $O_i = \{c\in \mathcal{C}:\ \bar{f}_i(c)=1\}$ as the set of all the sampled concept tuples that make true the $i$-th output explanation, i.e. the \emph{support} of $\bar{f}_i$. 


\paride{Example-level explanations.}
Given a sample $c\in O_i \subseteq \mathcal{C}$, the Booleanization $\sm{\bar{c}}$ of its continuous features may provide a natural way to get an example-level logic explanation $\varphi_{i,c}$. To make logic formulas more interpretable, the notation $\tilde{c}$ denotes human-interpretable strings representing the concept names or their negation,
\begin{equation}
    \varphi_{i,c} = \tilde{c}_1\wedge\ldots\wedge\tilde{c}_{m_i} \qquad\mbox{where } \tilde{c}_j:=
    \begin{cases}
        \bar{c}_j, & \mbox{if }c_j \geq 0.5\\
        \neg \bar{c}_j, &\mbox{if } c_j < 0.5\\
    \end{cases}\ ,
    \mbox{ for $j=1,\ldots,m_i$.}
    \label{eq:locexp}
\end{equation}

\paride{Set-level and Class-level explanations.}
By considering Eq. \ref{eq:locexp} for any $c\in S$, with $S\subseteq O_i$, and aggregating all the example-level explanations, an explanation for a set of samples can be generated as follows:
\begin{equation}
    \varphi_{i,S} = \displaystyle\bigvee_{c \in S}\varphi_{i,c} = \displaystyle\bigvee_{c \in S}\tilde{c}_1\wedge\ldots\wedge\tilde{c}_{m_i}
\label{eq:agg_exp}
\end{equation}
\draft{\sm{As some $\varphi_{i,c}$'s might be equivalent for different $c$'s, repeated instances can be discarded keeping only one of them, without loss of generality.} In case $S=O_i$, we simply write $\varphi_i$ in place of $\varphi_{i,S}$ and we refer to such set-level explanation as the \paridelist{class-level explanation} of the $i$-th output concept $f_i$. 

\sm{In the rest of this section we will explore further details of what has been described so far. We will start by the following example.}}


\begin{example}
\label{ex:xor}
Let's consider the Boolean $XOR$ function, defined by $xor(0,0)=xor(1,1)=0$, $xor(1,0)=xor(0,1)=1$.
Let $f = [f_1]$ be a LEN that has been trained to approximate the $XOR$ function in the input space $C=[0,1]^2$. Then, if we consider, e.g. the inputs $c^1=(0.2,0.7)$, $c^2=(0.6, 0.3)$, we get $\bar{c}^1=(0,1)$, $\bar{c}^2=(1,0)$, and therefore $\bar{f}_1(c^1)=\bar{f}_1(c^2)=1$. These examples yield the example-level explanations $\varphi_{1,c^1}=\neg \bar{c}_1\wedge \bar{c}_2$ and $\varphi_{1,c^2}=\bar{c}_1\wedge\neg \bar{c}_2$ respectively. As a result, the class-level explanation for $f_1$ is given by $\varphi_1=(\neg \bar{c}_1\wedge \bar{c}_2)\vee(\bar{c}_1\wedge\neg \bar{c}_2)$, which correctly matches the truth-table of the Boolean $XOR$ function. 


\end{example}

It is worth noting that both example and set-level explanations can be recursively applied in case of LENs with multiple $[0,1]$-valued hidden layers or in case of cascading LENs. For instance, given a cascading LEN, as the one in Fig.~\ref{fig:CascadeExplainer}, we may get both example and class-level explanations of the concepts in $E_1 = C_2$ with respect to the ones in $C_1$, so as the concepts in $E_2$ with respect to the ones in $C_2$ and, in turn, in $C_1$. The modularity of logic formulas allows the composition of categories at different levels that may express arbitrary complex relationships among concepts.

\draft{Logic explanations $\varphi_{i,S}$ generally hold only on a sub-portion $S$ of the sampled concept space $\mathcal{C}$.
However, we may get a logic formula providing an explanation holding everywhere by means of the disjunction
$\varphi = \varphi_1\vee\ldots\vee\varphi_r$, if we assume $O_1\cup\ldots\cup O_r=\mathcal{C}$.} Since $\varphi$ can turn out to be of little significance, if we are interested in simpler (and clearer) global explanations we may  convert $\varphi$ into an equivalent $\varphi'$ in \emph{Conjunctive Normal Form} (CNF). In particular, there always exist $r'$ and some clauses $\varphi'_1,\ldots,\varphi'_{r'}$ such that:
\begin{equation}
    \varphi=\bigvee_{i=1}^r\varphi_i \equiv \bigwedge_{i=1}^{r'}\varphi'_i=\varphi'.
\label{eq:dnf2cnf}
\end{equation}
As a result, we get a set of $r'$ explanations holding on the whole $\mathcal{C}$, indeed  $\varphi'_i(c)=1$ for every $c\in \mathcal{C}$, $i=1,\ldots,r'$.
Unfortunately, converting a Boolean formula from DNF into CNF can lead to an exponential explosion of the formula.
However, after having converted $\varphi_i$ in CNF, the conversion can be computed in polynomial time with respect to the number of minterms in $\varphi_i$ \cite{russell2016artificial}.

The methodologies described so far illustrate how logic-based explanations can be aggregated to produce a wide range of explanations, from the characterization of individual observations to formulas explaining model predictions for all the samples leading to the same output concept activation.
The formula for a whole class can be obtained by aggregating all the minterms corresponding to example-level explanations of all the observations having the same concept output.
In theory, this procedure may lead to overly long formulas as each minterm may increase the complexity of the explanation. In practice, we observe that many observations share the same logic explanation, hence their aggregation may not change the complexity of the class-level formula (right-side Fig. \ref{fig:tt2form_freq}). In general, ``\emph{satisficing}'' class-level explanations can be generated by aggregating the most frequent explanations for each output concept, avoiding a sort of ``explanation overfitting'' with the inclusion of noisy minterms which may correspond to outliers \cite{simon1956rational}.

As a final remark, a possible limitation of LENs can be the readability of logic rules. This may occur when ($i$) the number of input \draft{concepts} (the length of any minterm) $k \gg 1$, or ($ii$) the size of the support $|O_i|$ is very large (possibly getting too many different minterms for any $f_i$). In these scenarios, viable approaches to generate concise logic rules are needed to provide interpretable explanations. More details on how to generate concise explanations are in Section \ref{sec:parsimony}.

\subsection{Learning criteria}
\label{sec:learncrit}

In this section, we describe some of the loss functions allowing LENs to provide FOL explanations $\rho_i \in P$, according to the \sm{objectives} introduced in Section~\ref{sec:explanation_styles} (\paridelist{interpretable classification}, \paridelist{explaining a black-box} and \paridelist{interpretable clustering}). 
For simplicity, here we will not distinguish among the different computational pipelines, and we will refer to a generic LEN with $r$ output units.
\draft{We just saw how a logic rule $\varphi_i$ can be associated to an output concept $f_i$. However to improve the expressiveness of logic explanations, \sm{in this section we will promote $\varphi_i$ to a FOL formula $\rho_i$, depending on the selected learning criterion. This follows the usual approach adopted when a model combining logic and machine learning, trained on a finite collection of data, is then applied to out-of-sample inputs, following related studies \cite{ciravegna2020constraint,ciravegna2020human,gnecco2015foundations}}. As a result, any $\bar{c}_j$ \sm{that composes $\varphi_i$} is thought of as a logic predicate defined on the concept space $C$, and such that $\bar{c}_j(c)=1$ if and only if $c_j>0.5$, for any $c\in C$. In addition, if for instance $\varphi_i=\bar{c}_2\wedge\neg \bar{c}_5$, we will write $\varphi_i(c)$ for $\bar{c}_2(c)\wedge\neg \bar{c}_5(c)$.
Then, the specific choice on the loss function that drives the learning criteria of LENs introduces a link between $\varphi_i(c)$ and the final FOL formulas $\rho_i$ that is produced by the network.}

\paride{Interpretable classification.}
\draft{Supervised learning is needed to extract explanations for specific categories from LENs. This learning approach binds a logic explanation $\varphi_i$ to a specific output class of a classification problem. By denoting with $\bar{y}_i$ the binary predicate associated to the output class $i$, we will consider three kinds of FOL explanations $\rho_i$, between $\varphi_i$ and $\bar{y}_i$, expressed as the universal closure of an IF, Only IF or IFF rule.
These rules can be imposed in case of interpretable classification according to the following learning criteria.
}

IF rules mean that, in the extreme case, for each sample of class $i$ we want the $i$-th output of the LEN to score $1$ (but not necessarily the opposite). In other words, \draft{the set of concept tuples $c$ belonging to the $i$-th class, i.e. such that $\bar{y}_i(c)=1$ has to be included in the support of $\bar{f}_i$, \sm{while no conditions are imposed when $\bar{y}_i(c)=0$ (recall that $f_i(c) \in [0,1]$)}.}
This behavior can be achieved by minimizing a hinge loss,
\begin{equation}
L_{\rightarrow}(\bar{y}_i,f_i,\mathcal{C}) = \sum_{c\in \mathcal{C}} \max\{0, \bar{y}_i(c) - f_i(c)\}\qquad  i \in [1,  r].\label{eq:IF}
\end{equation}
 Following a symmetric approach, in Only IF rules a class is explained in terms of lower-level concepts. This principle is enforced by swapping the two terms in the loss function in Eq. \ref{eq:IF},
\begin{equation}
L_{\leftarrow}(\bar{y}_i,f_i,\mathcal{C}) = \sum_{c\in \mathcal{C}} \max\{0, f_i(c) - \bar{y}_i(c)\}\qquad  i \in [1,  r].\label{eq:IF2}
\end{equation}
Both in IF and Only IF rules, further conditions on $f$ must be included in order to make the learning problem well posed. To this aim, we need to define the behavior of the model in regions of the concept space not covered by training samples in order to avoid trivial solutions with constant $f$. \sm{For example, $f$ could have a fixed bias equal to $1$ or no biases at all.}

At last, LENs can learn double implication rules (IFF) which completely characterize a certain class. \sm{Our experiments are focused on this type of explanations}. 
Any function penalizing points for which $f_i(c)\neq \bar{y}_i(c)$ may be employed in this scenario, such as the the classic Cross-Entropy loss,
\begin{equation}
L_{\leftrightarrow}(\bar{y}_i,f_i,\mathcal{C}) = \sum_{c\in \mathcal{C}} \bar{y}_i(c)\log(f_i(c)) + \left( 1-\bar{y}_i(c) \right) \log \left( 1-(f_i(c)) \right).
\label{eq:IFF}
\end{equation}
\draft{For all the above loss functions (Eq.~\ref{eq:IF}-\ref{eq:IFF}), LENs provide logic explanations in First-Order Logic by means of the following equations:
\begin{align}
\text{IF-rule}: \qquad &  \rho_i = \forall c\in C:\ \bar{y}_i(c)\rightarrow\varphi_i(c). \label{eq:FOLexplIF}\\ 
     \text{Only IF-rule}: \qquad  & \rho_i = \forall c\in C:\ \varphi_i(c) \rightarrow  \bar{y}_i(c) \label{eq:FOLexplFI}\\
     \text{IFF-rule}: \qquad & \rho_i = \forall c\in C:\  \bar{y}_i(c) \leftrightarrow \varphi_i(c)\label{eq:FOLexplIFF}
\end{align}
where each $\rho_i$, for $i=1,\ldots,r$ corresponds to a FOL formula, ranging on the whole concept space $C$, \sm{thus generalizing the relationships discovered on the data samples}.
In the same way, in case of a concept-bottleneck pipeline, a FOL explanation can be derived from the function $g:X\to C$ extracting LEN's input concepts from raw features. For instance for the IFF-rule, we will get
\begin{equation}
\rho_i=\forall x \in  X:\ \bar{y}_i(g(x))\leftrightarrow\varphi_i(g(x))
\label{eq:FOLexplCB}
\end{equation}
\sm{We remark that} the logic predicate $\bar{y}_i$ appearing in the Eq.~\ref{eq:FOLexplIF}-\ref{eq:FOLexplCB}
is simply a \textit{virtual} predicate denoting the membership of a certain concept tuple $c\in C$ to the $i$-th output class. 
%
Despite the correspondence between $f_i$ and $\bar{y}_i$ can be enforced only on the sampled concept  space $\mathcal{C}$, we assume that any $\rho_i$ can generalize to unseen concept tuples in the whole concept space $C$, in line with previous works \cite{ciravegna2020human,ciravegna2020human,gnecco2015foundations}.
}


\paride{Explaining a black-box.} In this case, LEN's outputs are forced to mimic the predictions of a black-box \draft{$g:C\rightarrow Y$, instead of ground-truth labels. This behaviour can be imposed by considering loss functions analogous to the ones in Eq.~\ref{eq:IF}-\ref{eq:IFF}, with $g_i$ in place of $\bar{y}_i$. For instance, for the IFF rule}
the coherence loss \sm{of Eq.~\ref{eq:IFF}} can be leveraged, \sm{allowing LENs to mimic the behaviour of the black-box},
\begin{equation}
L_{\leftrightarrow}(g_i,f_i,\mathcal{C}) = \sum_{c\in \mathcal{C}} g_i(c)\log(f_i(c)) + \left( 1-g_i(c) \right) \log \left( 1-(f_i(c)) \right)
 \label{eq:coher}
\end{equation}
\draft{As in the case of interpretable classification, IF, Only IF and IFF rules can be expressed according to Eq.~\ref{eq:FOLexplIF}-\ref{eq:FOLexplCB}. However, here FOL explanations will hold assuming that $\bar{y}_i(c)$ \sm{is not a virtual predicate, but it is explicitly associated to the (Booleanized) black-box predictions $g_i(c)$.}
}
%


\paride{Interpretable clustering.}
Generic explanations can be obtained by means of fully unsupervised  principles \sm{we borrow} from Information Theory. As a matter of fact, the maximization of the Mutual Information (MI) index between the input concept space $C$ and the output concept space $E$ allows LENs to
be trained in a fully unsupervised way \cite{ciravegna2020constraint}. More specifically, a max-MI criterion (see \cite{melacci2012unsupervised} for further details) leads to LENs leaning towards 1-hot activation scores, such that $\forall c \in \mathcal{C}$ only one $f_i(c) \simeq 1$, while the others are close to zero. This encourages LENs to cluster input data such that each input sample belongs to a single cluster.
In order to define the MI index, we have to model the probability distribution of each $f_i$ to be active (close to 1) on a given sample $c$, that we implemented using the softmax operator on LENs' outputs. 
The learning criterion to minimize in order to train LENs is minus the MI index,
\begin{equation}
    L_{MI}(f,\mathcal{C}) = - H_{E}(f,\mathcal{C}) + H_{E|C}(f,\mathcal{C}) \ ,
	\label{eq:MI}
\end{equation}
where $ H_{E}$ and $ H_{E|C} $ denote  the entropy and the conditional entropy functions associated to the aforementioned probability distribution, respectively, and measured over the whole $\mathcal{C}$, as described by Ciravegna et al.  \cite{ciravegna2020constraint}.
\draft{
In this case, the support $O_i$ of $\bar{f}_i$ is exactly the cluster of data points where the $i$-th output of the LEN is active. Leaving $\varphi_i$ free to relate concepts in a purely unsupervised manner, we naturally get the FOL explanation}
\begin{equation}
\rho_i=\forall c \in O_i:\ \varphi_i(c).
\label{eq:FOLexplMI}
\end{equation}

As a final remark, in all the computational pipelines in which a classifier $g$ is employed,
the available supervision is enforced on $g$ as well, e.g. by means of the Cross-Entropy loss. As a side note, we mention that what we are describing in a fully supervised setting can be trivially extended to the semi-supervised one, as investigated in previous works \cite{ciravegna2020constraint,ciravegna2020human}.

\subsection{Parsimony} \label{sec:parsimony}
When humans compare a set of explanations outlining the same outcomes, they tend to have an implicit bias towards the simplest one \cite{aristotlePosterior,mackay2003information}. 
\sm{In the case of LENs, the notion of simplicity is implemented by reducing the dependency of each output unit by all of the $k$ input concepts, encouraging only a subset of them to have a major role in computing LENs' outputs. Such subset is of size $m_i \leq k$ for the $i$-th output unit.}

Over the years, researchers have proposed many approaches to integrate ``\textit{the law of parsimony}'' into learning machines. \sm{These approaches can be eventually considered as potential solutions to implement parsimony criteria in LENs, in order to find a valid way to fulfill the end-user requirements on the quality of the explanations and on the classification performance.}
\sm{For instance,} Bayesian priors \cite{wilson2020case} and weight regularization \cite{kukavcka2017regularization} are two of the most famous techniques to put in practice the Occam's razor principle in the fields of statistics and machine learning. 
Among such techniques, $L1$-regularization \cite{santosa1986linear} has been recently shown to be quite effective for neural networks providing logic-based explanations \cite{ciravegna2020human} as it encourages weight sparsity 
by shrinking the less important weights 
to zero  \cite{tibshirani1996regression}. 
 \sm{This allows the model to ignore some of the input neurons, that, in the case of the first layer of a LEN, corresponds to discarding or giving negligible weight to some of the input concepts, opening to simplified FOL explanations.
If $W$ collects (a subset of) the weights of the LEN that are subject to this regularization, the learning criterion is then augmented by adding $\lambda \| W \|_1$. Notice that the precise weights involved in $W$ might vary in function of the selected neural architecture to implement the LEN, that is the subject of Section~\ref{sec:outofthebox}.}
All the out-of-the-box LENs of the next section are trained using this parsimony criterion, \sm{that acts in different portions of the network in the considered instances of LENs. The parsimony criterion is usually combined with a pruning strategy amongst the ones that are defined in the following.} 

\subsubsection{Pruning strategies}
\label{sec:pruning}
The action of parsimony criteria, such as regularizers or human priors, influences the learning process 
towards specific local minima. Once the model has finally converged to the desired region of the optimization space, the effort can be speed up and finalized by pruning the neural network \cite{lecun1989optimal,hassibi1993second} i.e., \sm{removing connections whose likelihood of carrying important information is low. The choice of the connections to be pruned depends on the selected pruning strategy}. Such a strategy has a profound impact both on the quality of the explanations but also on the classification performance \cite{frankle2018lottery}. Here we present three effective pruning strategies specifically devised for LENs, \sm{whose main goal is to keep FOL formulas compact for each LEN's output, namely \paridelist{node-level pruning}, \paridelist{network-level pruning}, \paridelist{example-level pruning}.}

\paride{Node-level pruning.}
The most ``fine-grained'' pruning strategy considers each neuron of the network independently.
This strategy requires the user to define in advance the maximum \textit{fan-in} $\zeta \in \mathbb{Z}^+$ for each neuron of a feed-forward neural network i.e., the number of non-pruned incoming connections each neuron can support. \sm{In this case, the pruning strategy removes all the connections associated to the smallest weights entering the neuron, until the target fan-in is matched}. We refer to this strategy as \textit{node-level pruning}.
\sm{In detail,} the node-level approach prunes one by one the weights with the smallest absolute value, such that each neuron in the network has a fixed number \sm{$\zeta$} of incoming non-pruned weights. The parameter \sm{$\zeta$} determines the computational capabilities of the pruned model as well as the complexity of the logic formulas which can be extracted, \sm{as it reduces the number of paths connecting the network inputs to each output neuron}. To get simple explanations out of each neuron, the parameter $\zeta$ may range between $2$ and $9$ \cite{miller1956magical,cowan2001magical,ma2014changing}). Recent work on explainer networks has shown how node-level pruning strategies may lead to fully explainable models  \cite{ciravegna2020constraint}. However, we will show in the experimental section that this pruning strategy strongly reduces the classification performances.

\paride{Network-level pruning.}
In order to overcome the heavy reduction in \sm{learning} capacity of node-level pruned models, we introduce the so-called \textit{network-level pruning}.
\sm{This pruning operation aims at reducing the number of concepts involved in FOL explanations by limiting the availability of input concepts.} 
\sm{In detail, the $L$2 norm of the connections departing from each input concept is computed.} \sm{If $w = [w_1,\ldots,w_k]$ is the vector that collects the resulting $k$ norms}, then we re-scale it in the interval $[0,1]$,
\begin{equation}
    w' = \frac{w}{\max_j \{ w_j:\ j=1,\ldots,k \}}
\end{equation}
\sm{where the division is intended to be applied in an element-wise fashion.}
In this way, $w'$ gives a normalized score to rank input concepts by importance. The \textit{network-level strategy} consists in pruning all the input features for which $w_j' < \tau$, where $\tau$ is a custom threshold (in our experiments, $\tau = 0.5$). Alternatively, 
we can retain the \sm{$\zeta$} most important input concepts discard all the others. This can be achieved by pruning all the connections departing from the least relevant concepts similarly to \textit{node-level pruning}. 
Anyway, this pruning strategy is far less stringent compared to node-level pruning as it affects only the first layer of the network and it does not prescribe a fixed fan-in for each neuron.

\paride{Example-level pruning.}
Example-level pruning is a particular strategy leveraging the Voronoi tessellation generated by neural networks whose activation functions in all hidden layers are Rectified Linear Units (ReLU Networks) \cite{hahnloser2000digital}.
\sm{If the LEN is implemented as a ReLU Network}, for any input $c \in C$, the Directed Acyclic Graph (DAG) ${\cal G}$ describing the structure of the connections in the LEN, can be reduced to ${\cal G}_{c}$,
which only keeps the units corresponding to active neurons (the ones for which the ReLU activation is non-zero) and the corresponding arcs \sm{(referred to as the ``firing path'')}.
Since all neurons operate in ``linear regime'' (affine functions) \sm{in the reduced ${\cal G}_{c}$}, as stated in the following,
\sm{the input-to-output transformation computed by the multi-layer feed-forward ReLU network \sm{with structure ${\cal G}_{c}$} is a composition of affine functions over the network hidden layers that, in turn, can be simplified with a single affine function, leading to 
\begin{equation}
    f(c) = \sigma (\hat{W}^{(c)} c + \hat{b}^{(c)}), 
    \label{eq:single}
\end{equation}    
\sm{being $\sigma$ the activation of the output layer and $\hat{W}^{(c)}$, $\hat{b}^{(c)}$ a weight matrix and biases (respectively) computed as described in the following}.}
\begin{theorem}
	\sm{Let $\{\xi_1,\ldots,\xi_L \}$ be a collection of affine functions, where
		$\xi_{\kappa}: U_{\kappa} \mapsto V_{\kappa}: u \mapsto W_{\kappa} u + b_{\kappa}$
	and $\forall \kappa=1,\ldots, L-1: \ \ U_{\kappa+1} \subset V_{\kappa}$. If a multi-layer network computes the last layer activations as
		$\xi_{L} \circ \xi_{L-1} \circ \dots \circ \xi_{2} \circ \xi_{1}$,
	then such transformation is affine and we can re-write it as $\hat{W} c + \hat{b}$, where
	\begin{equation}
	    \hat{W} = \prod_{\kappa=1}^{L} W_{\kappa}, \qquad
	    \hat{b} = \sum_{\kappa=0}^{L-1}  b_{L-\kappa} \prod_{h=0}^{\kappa} W_{L-h+1}
	\end{equation}
	and $W_{L+1}:=I$.}
	\label{theorem:affine}
\end{theorem}

\sm{The proof of Theorem \ref{theorem:affine} is straightforward. Such a theorem perfectly applies to the case of ReLU networks once $\mathcal{G}_c$ has been fixed, since they boil down to simple networks with linear activations. The theorem does not introduce any conditions on how input is represented or on the activation functions in the output layer.} 
\sm{Given $c \in C$, once \sm{${\cal G}_{c}$} has been determined, the function computed by the (deep) LEN has the following form, $f(c) = \sigma( \xi_{L}^{(c)} \circ \xi_{L-1}^{(c)} \circ \dots \circ \xi_{2}^{(c)} \circ \xi_{1}^{(c)} (c) )$, that reduces to $f(c) = \sigma (\hat{W}^{(c)} c + \hat{b}^{(c)})$, where the superscript ${(c)}$ is added to the symbols of Theorem \ref{theorem:affine} to highlight the fact they are specifically instantiated in the case of the network with structure ${\cal G}_{c}$. The reduced form of $f$ is much easier than considering the one of the original deep network.
However, ${\cal G}_{c}$ might vary for each input $c$, thus we might get different transformations for different input samples.}
Indeed, \sm{for each sample we can compute the corresponding ${\cal G}_{c}$ and, in turn, $\hat{W}^{(c)}$ and $\hat{b}^{(c)}$, that we can prune as described in the case of network-level pruning, keeping only the connections associated with the most important concepts (for that specific sample).}

\section{Out-of-the-box LENs}\label{sec:outofthebox}

\begin{figure}[!t]
    \centering

     \begin{subfigure}[t]{0.55\textwidth}
         \centering
         \includegraphics[width=1.0\textwidth]{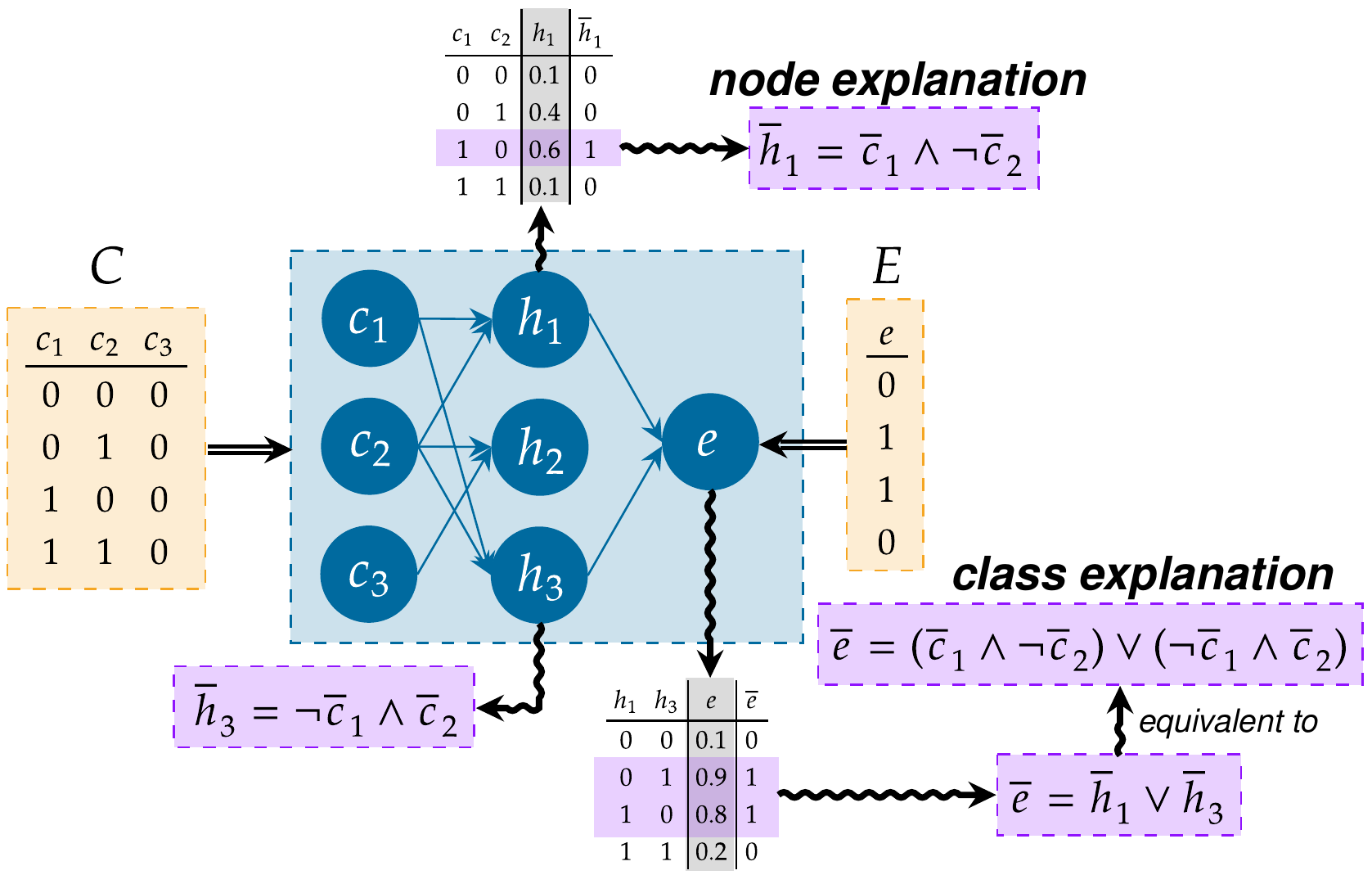}
         \caption{$\psi$ network (Section \ref{subsec:psi}).}
         \label{fig:psi_net}
     \end{subfigure}
     \begin{subfigure}[t]{0.40\textwidth}
         \centering
         \raisebox{10mm}{\includegraphics[width=1\textwidth]{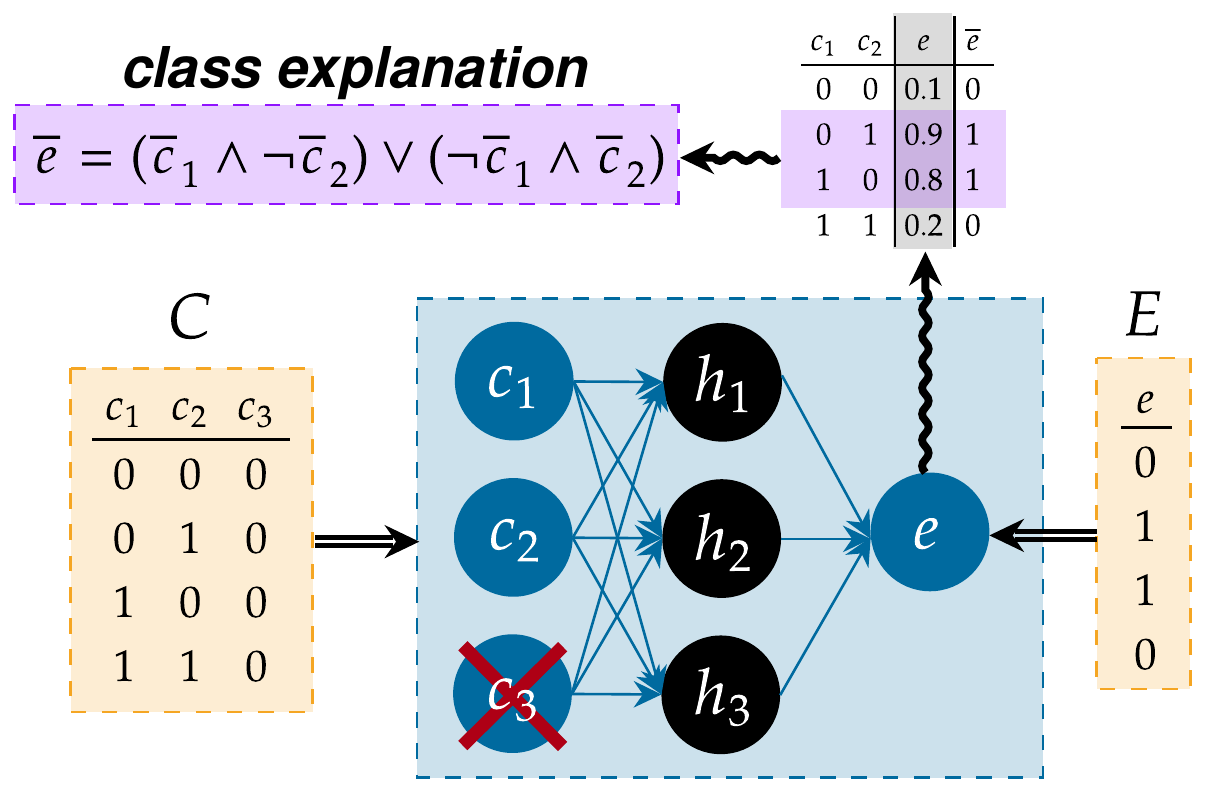}}
         \caption{$\mu$ network (Section \ref{subsec:mu}).}
         \label{fig:mu_net}
     \end{subfigure}\\
  
     \begin{subfigure}[t]{0.37\textwidth}
         \centering
         \includegraphics[width=1\textwidth]{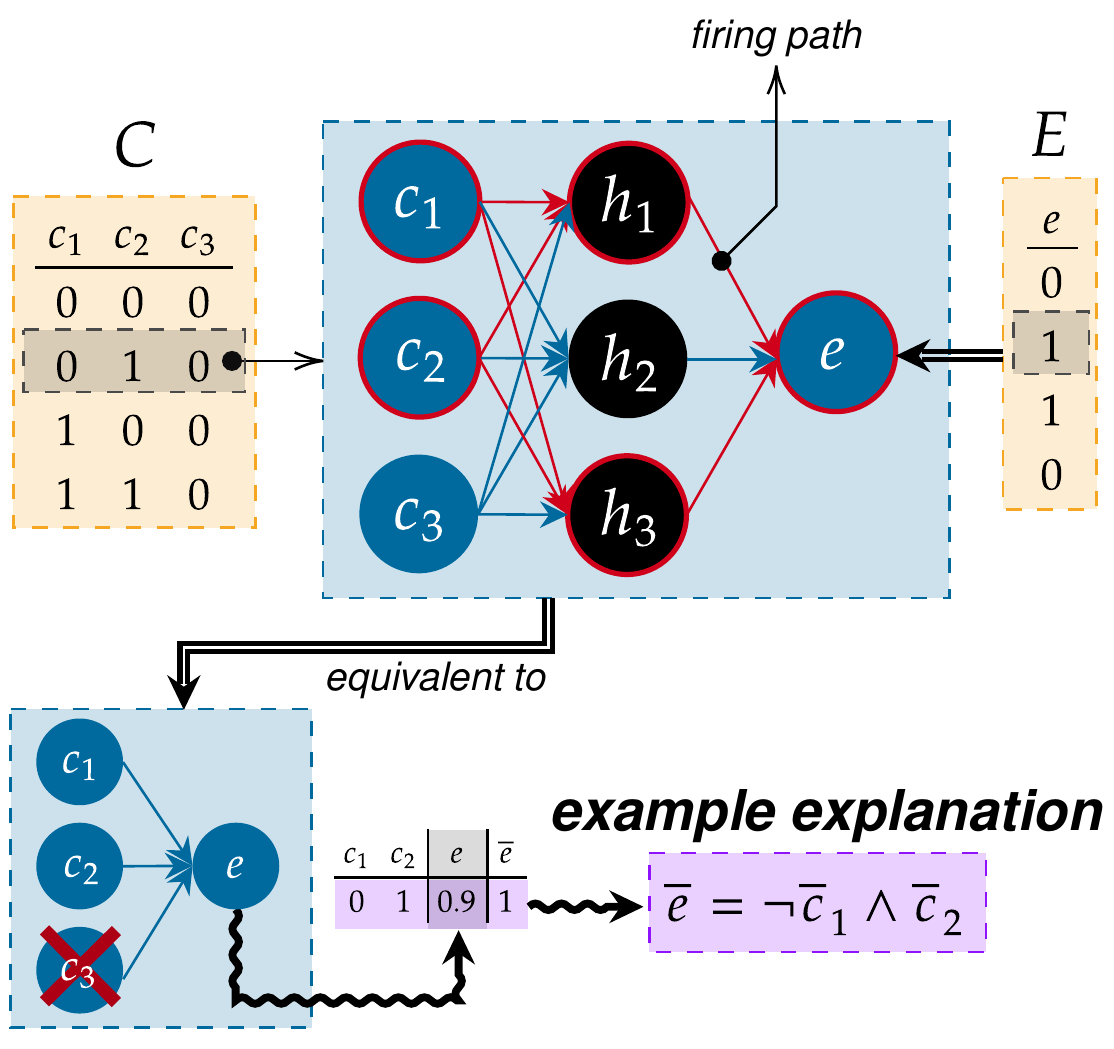}
         \caption{ReLU network (Section \ref{subsec:ReLU}).}
         \label{fig:ReLU_net}
     \end{subfigure}

    \caption{Out-of-the-box LENs  \sm{showcased in the case of interpretable classification, with examples of logic rules extracted using the procedure of  Section~\ref{sec:logic_rule_extr}}. (a) $\psi$ networks
    only admit $[0,1]$-valued neurons with low fan-in, hence we can associate a Boolean formula to each neuron by analysing its I/O truth-Table~(b) $\mu$ networks
    do not have interpretable hidden neurons but the input concepts are drastically pruned which allows to provide simple network-level explanations. (c) ReLU networks
    supply explanations for each example by means of the equivalent affine transformation. However, nor the nodes, nor the network can be logically interpreted.}
    \label{fig:out-of-the-box}
\end{figure}

Crafting state-of-the-art fully interpretable models is not an easy task; rather, there is a trade-off between interpretability and performances.
The framework \sm{introduced in Section~\ref{sec:basics}, with the methods described in Section~\ref{sec:methods}}, is designed to provide the building-blocks to create a wide range of models having different interpretability vs. accuracy
trade-offs.
Here, we \sm{showcase three out-of-the-box neural networks implementing different LENs, whose key properties are about different ways of leveraging the parsimony strategies, as visually anticipated in Fig.~\ref{fig:visual} (right). In Fig. \ref{fig:out-of-the-box} we sketch the so-called $\psi$, $\mu$, and ReLU out-of-the-box LENs that will be described in the following. Briefly,} the $\psi$ network, originally proposed in \cite{ciravegna2020constraint}, is a fully interpretable model with limited learning capacity providing mediocre explanations; the $\mu$ network is a neural model that can provide high-quality explanations, good learning capacity and modest interpretability; the ReLU network is a model enabling state-of-the-art learning capabilities and good explanations at the cost of very low interpretability. The characteristics of these three LENs are summarized in Table~\ref{tab:LENs}.

\begin{table}[ht]
\centering
\caption{Out-of-the-box LENs and their main properties.}
\label{tab:LENs}
\resizebox{1.0\textwidth}{!}{
\begin{tabular}{llllll}
\toprule
\textbf{LEN}                            & \textbf{Pruning} & \textbf{Activation}  & \textbf{Learning}  & \textbf{Explanation}   & \textbf{Interpretability}\\
\midrule

$\psi$ Net (Fig. \ref{fig:psi_net}) & Node-level        & Sigmoid                       & Low                           & Low                               & Very high     \\
$\mu$ Net (Fig. \ref{fig:mu_net})   & Network-level     & Any                           & High                          & Very high                         & High          \\
ReLU Net (Fig. \ref{fig:ReLU_net})  & Example-level     & ReLU                          & Very high                     & High                              & Low           \\

\bottomrule
\end{tabular}
}
\end{table}

\subsection{$\psi$ Network} \label{subsec:psi}
A $\psi$ network, originally proposed in \cite{ciravegna2020constraint,ciravegna2020human}, \sm{is the first model in the LENs family}. A $\psi$ network is based on three design principles:
($i$) all activation functions for all neurons (including hidden units) should be sigmoids;
($ii$) a strong $L$1-regularization is used in all the layers to shrink the less important weight values towards zero;
($iii$) a node-level pruning strategy is considered, where each neuron of the network must have the same number of incoming non-pruned weights (suggested values are between 2 and 9). This number is directly proportional to the number of terms involved in the explanations.
\sm{In line with previous works \cite{ciravegna2020constraint,ciravegna2020human}, the rule generation mechanism involves the extraction of rules from each neuron of the network (also the hidden ones), and the node-level pruning favours simple rules on each neuron. Rules are then combined to get the final explanations. The number of hidden layers in the network needs to be small to avoid issues in the rule-merging procedure. In our implementation, pruning is performed after half of the training epochs has been completed, so that the pruned network can refine the value of the remaining weights during the last epochs.}

\sm{We can cast the $\psi$ network as a member of the LENs family.} The computational pipeline presented in the original work corresponds to a Concept-Bottleneck pipeline.
\sm{However, from the perspective of the rule-extraction mechanism, the $\psi$ network can be seen as a Cascading LEN where each layer corresponds to a new level of LEN, each of them generating rules. However, the network is trained as a whole, since only the first layer gets human-understandable concepts. 
Neuron-specific explanations are then aggregated in order generate final explanations involving input concepts only.}
For instance, by recalling Example \ref{ex:xor}, a $\psi$ network with \sm{a single hidden layer with} two hidden nodes may learn the logic formula:
\begin{equation}
\varphi_1= (\neg \bar{c}_1\wedge \bar{c}_2)\vee(\bar{c}_1\wedge\neg \bar{c}_2),
\label{eq:xor}
\end{equation}
with hidden nodes learning the formulas $\bar{h}_1= \neg \bar{c}_1\wedge \bar{c}_2$ and $\bar{h}_2 = \bar{c}_1\wedge \neg\bar{c}_2$ respectively, and the output node $f_1$ learning $\varphi_1 = \bar{h}_1 \vee \bar{h}_2$.

We remark that the $\psi$ network is specifically designed to be a fully interpretable neural network. 
As shown in Fig. \ref{fig:psi_net}, this network provides a high level of interpretability as each neuron can be explained.
On the contrary, the strong regularization and the restrictive pruning strategy lead to poor classification accuracy making $\psi$ networks hard to train and not suitable for solving complex categorical learning problems. Besides, the provided explanations may be disappointing, due to the \sm{limited learning capabilities of the model}.

\subsection{$\mu$ Network}  
\label{subsec:mu}
A $\mu$ network is a LEN based on a multi-layer perceptron \sm{without the strong constraints on neurons' fan-in of the $\psi$ net. In particular,} a $\mu$ network is based on two design principles:
    ($i$) a strong L1-regularization is used \sm{in the weights connecting the input to the first hidden layer of the network};
    ($ii$) a network-level pruning strategy, which prunes input neurons only.
After the pruning step, the $\mu$ network can be fine-tuned to \sm{better adapt it} to the new configuration (we pruned the network after half of the training epochs has completed). No assumptions need to be made nor on hidden layers nor on activation functions. However, by applying a network-level pruning strategy, the same set of retained \sm{input concepts} is used for the whole network. On one hand, this allows a simple logic interpretation of the network and, therefore, to provide concise explanations. On the other hand, this may represent a severe limitation for multi-class problems as each class may rely on \sm{its own input concepts}, but these may be very different among the classes. \sm{However, $\mu$ networks can be efficiently adapted to multi-class experiments (as the ones of Section \ref{sec:experiments}) by splitting the multi-class problem into a set of binary classification problems (one per class), i.e. using a set of light binary $\mu$ networks.}

Still considering Example \ref{ex:xor}, $\mu$ networks with a  different number of hidden layers may easily learn the same logic formulas.
The network-level pruning strategy in this specific example should not prune any of the input features as both $c_1$ and $c_2$ are relevant for the target class. However, if we assume the presence of additional redundant features, they would be likely discarded in favor of $c_1$ and $c_2$, as shown in Fig. \ref{fig:mu_net}.
\sm{Assuming a successful training, the support set $O_1$ will be composed by the second and third examples of the yellow-box in Fig. \ref{fig:mu_net}, that will yield the following example-level explanations respectively}
\[
\neg c_1 \wedge c_2 \quad \text{and} \quad c_1 \wedge \neg c_2.
\]
The two explanations can be then considered altogether \sm{(Eq.~\ref{eq:agg_exp})} to explain the whole class $1$ (class-level explanation) by
\[
\varphi_1 = \bigvee_{c \in O_1}\varphi_{1,c} =
 (\neg \bar{c}_1\wedge \bar{c}_2) \vee ( \bar{c}_1\wedge \neg\bar{c}_2).
\]
Thanks to the permissive pruning strategy, the learning \sm{capabilities} of the network \sm{almost match an unconstrained network}. As a consequence, $\mu$ networks are suitable for solving and explaining more complex categorical learning problems. \sm{As we will show in Section~\ref{sec:experiments},} the quality of the explanations provided by the $\mu$ networks are among the highest of the proposed models. At last, a mild-level of interpretability is guaranteed as $\mu$ nets can be logically interpreted as a \sm{whole, but hidden neurons are not as interpretable as in $\psi$ networks}.

\subsection{ReLU network}  \label{subsec:ReLU}
The ReLU network is another member of the LENs family providing a different accuracy vs. interpretability trade-off. 
This model is based on three design principles:
    ($i$) all activation functions for all \emph{hidden} neurons are rectified linear units;
    ($ii$) a mild L1-regularization is applied \sm{to all the weights associated to each layer of the network};
    ($iii$) an example-level pruning strategy is used, i.e. a specialized pruning strategy for each sample.
\sm{Principle ($iii$) can be applied due to the presence of} rectified linear units activation functions. 
The restriction to ReLU activations is not as limiting as it sounds since it is among the most widely used and efficient activation functions employed in deep learning \cite{glorot2011deep,ramachandran2017searching}.
\sm{What makes this LEN significantly different from both $\psi$ and $\mu$ nets, is that the pruning strategy does not alter the network structure at all. This is due to the fact that pruning is applied to the weights  that belong to the single-affine instance of $f(c)$ of Eq.~\ref{eq:single}, whose values are collected in $\hat{W}^{(c)}$ and are only computed for rule-extraction purposes.} This means that the original capacity of the model is fully preserved, eventually leading to state-of-the-art classification performances. However, this type of pruning does not provide general insights about the model behaviour, \sm{as it is only about the considered example $c$}, and they may not always lead to optimal explanations.

Recalling again Example \ref{ex:xor}, a ReLU network with  hidden layers \sm{can learn the correct logic formula by aggregating different example-level explanations, as we did in the case of the $\mu$ network. For example, in Fig.~\ref{fig:ReLU_net} we show the case of the processing the second sample from the yellow table, that provides a portion of the explanation of the XOR function.} \sm{However, when we restrict the connections to the arcs of $\mathcal{G}_{c}$ for a certain sample $c$, we actually discard several weights of the original ReLU network, i.e. the ones of all the connections that are not needed for classifying the considered $c$ correctly. This implies that the single-affine instance of $f(c)$ of Eq.~\ref{eq:single}, being it function of $\mathcal{G}_c$, has a very localized dependence on the space region to which $c$ belongs. Since Eq.~\ref{eq:single} is the form of the LEN from which we extract the example-level explanation, there is the serious risk of obtaining an explanation that does not carry much information from the original structure $\mathcal{G}$ and that does not globally applies to the whole class. As a matter of fact, there might be strong differences on the example level explanations of samples, even when belonging to the same class. }

Summing up, this LEN has the capacity to provide the best performances in terms of classification accuracy thanks to the example-level pruning strategy \sm{that do not alter the original network}. However, this comes at the cost of poor model interpretability and mild explanation capacity.


\section{Benchmarking out-of-the-box LENs} \label{sec:experiments}
In this section, we quantitatively assess the quality of the explanations and the performance of LENs, compared to state-of-the-art white-box models. \sm{We consider several tasks, covering multiple combinations  of computational pipelines (Section~\ref{sec:use_cases}) and objectives (Section~\ref{sec:explanation_styles})}. The summary of each task is reported in Table \ref{tab:datasets}.
A brief description of each task, the corresponding dataset and all the related experimental details are exposed in Section \ref{sec:exp_details}. In Section \ref{sec:exp_metrics} six quantitative metrics are defined and used to compare LENs with the considered state-of-the-art methods.

\begin{table}[ht]
\centering
\caption{Summary of the experiments.}
\label{tab:datasets}
\resizebox{1.0\textwidth}{!}{
\begin{tabular}{llll}
\toprule
\textbf{Dataset} & \textbf{Description} & \textbf{Pipeline}  & \textbf{Objective}\\
\midrule

MIMIC-II  (Fig. \ref{fig:E2EExplainer}, top)                      & \small{Predict patient survival from clinical data}       &E2E & Interpretable classification   \\
MNIST E/O (Fig. \ref{fig:CMBExplainer}, top)                         & \small{Predict parity from digit images}                  & CB & Interpretable classification  \\
CUB (Fig. \ref{fig:len})                                    & \small{Predict bird species from bird images}             & CB & Interpretable classification \\
V-Dem (Fig. \ref{fig:CascadeExplainer})                     & \small{Predict electoral democracy from social indexes}   & Cascading & Interpretable classification  \\
MIMIC-II (EBB)     & \small{Predict patient survival from clinical data}       & E2E & Explaining black-box (Fig.~\ref{fig:E2EExplainer}, bottom)       \\
MNIST E/O (ICLU) (Fig. \ref{fig:CMBExplainer}, bottom)                     & \small{Cluster digit properties from digit images}        & CB & Interpretable clustering      \\
\bottomrule
\end{tabular}
}
\end{table}

The Python code and the scripts used for the experiments, including parameter values and documentation, is freely available under Apache 2.0 Public License from a GitHub repository\footnote{\url{https://github.com/pietrobarbiero/logic_explainer_networks}}. \sm{The code is based on our} "Logic Explained Networks" library \cite{barbiero2021lens}, designed to make out-of-the-box LENs accessible to researchers and neophytes by means of intuitive APIs requiring only a few lines of code to train and get explanations from a LEN, as we sketch in the code example of Listing \ref{code:example}.
\begin{center}
\begin{minipage}{0.8\textwidth}
\begin{lstlisting}[basicstyle=\ttfamily\tiny, language=Python, label=code:example, caption=Example on how to use the ``Logic Explained Networks'' library.]
import lens

# import train, validation and test data loaders
[...]

# instantiate a "psi network"
model = lens.models.PsiNN(n_classes=n_classes, n_features=n_features,
                       hidden_neurons=[200], loss=torch.nn.CrossEntropyLoss(),
                       l1_weight=0.001, fan_in=10)

# fit the model
model.fit(train_data, val_data, epochs=100, l_r=0.0001)

# get predictions on test samples
outputs, labels = model.predict(test_data)

# get first-order logic explanations for a specific target class
target_class = 1
formula = model.get_global_explanation(x_val, y_val, target_class)

# compute explanation accuracy
accuracy = lens.logic.test_explanation(formula, target_class, x_test, y_test)
\end{lstlisting}
\end{minipage}
\end{center}
Further details about low-level APIs can be found in Appendix \ref{sec:apis}.


\subsection{Dataset and classification task details}
\label{sec:exp_details}
We considered five categorical learning problems ranging from computer vision to medicine and democracy. Some datasets (e.g. CUB) were already annotated with high-level concepts (e.g. bird attributes) which can be leveraged to train a concept bottleneck pipeline. For datasets without annotations for high-level concepts, we transformed the input space into a predicate space (i.e. $\mathbb{R}^9 \rightarrow [0,1]^d$) to make it suitable for training LENs. According to the considered data type, we will evaluate and discuss the usage of  different LEN pipelines and objectives, as already anticipated in Table~\ref{tab:datasets}.
For the sake of consistency, \sm{in all the experiments based on supervised learning criteria (Section~\ref{sec:learncrit})} LENs were trained by optimizing Eq.~\ref{eq:IFF}, therefore extracting IFF rules, \sm{that better cope with the ground truth of the datasets}. However, as already discussed in Section~\ref{sec:learncrit}, also Eq.~\ref{eq:IF} and Eq.~\ref{eq:IF2} \sm{could be considered, yielding different target rule types. In the following we describe the details of each task.}

\paride{MIMIC-II - {\scriptsize E2E, Interpretable classification}.}
The Multiparameter Intelligent Monitoring in Intensive Care II (MIMIC-II, \cite{saeed2011multiparameter,goldberger2000physiobank}) is a public-access intensive care unit (ICU) database consisting of 32,536 subjects (with 40,426 ICU admissions) admitted to
ICUs at a single tertiary care hospital. The dataset contains detailed description of a variety of clinical data classes: general,
physiological, 
results of clinical laboratory tests, 
records of medications, fluid balance, 
and free text notes and reports of imaging studies (e.g. x-ray, CT, MRI, etc). In our experiments, we removed the text-based input features and
we discretized blood pressure (BP) into five different categories (one-hot encoded): very low BP, low BP, normal BP, high BP, and very high BP. After such preprocessing step, we obtained an input space $X$ composed of $90$ key features.
The task consists in training a classifier function to identify recovering or dying patients, after 28 days from ICU admission ($Y = [0,1]^{2}$). In particular, we considered the case of
\textit{interpretable classification} where an End-to-End LEN $C \rightarrow E$ is employed to \sm{directly} carry out the classification task, i.e. with $C=X$ and $E=Y$.

\paride{MIMIC-II (EBB) - {\scriptsize E2E, Explaining a black-box model}.} On the same dataset, a second task is set up. A black-box model $g$ is employed to solve the \sm{previously described} classification task. The end-to-end LEN $f$ instead is trained to mimic the behaviour of the black-box $g$, \sm{using the learning criterion of Eq.~\ref{eq:coher}.}


\paride{MNIST E/O - {\scriptsize CB, Interpretable classification}.}
The Modified National Institute of Standards and Technology database (MNIST, \cite{lecun1998mnist}) contains a large collection of images representing handwritten digits. The input space $X \subset \mathbb{R}^{28\times 28}$ is composed of 28x28 pixel images of digits \sm{from $0$ to $9$. However, the task we aim to solve is slightly different from the common digit-classification. We are interested in determining if a digit is either odd or even, and explain the assignment to one of these classes in terms of the specific digit categories from $0$ to $9$. In the notation of this paper, we can consider that each image comes with $12$ attributes, where the first $10$ ones are binary attributes about the specific digit type, i.e. from $0$ to $9$, while the last $2$ ones are binary labels that encode whether the digit is even or odd \sm{($Y=[0,1]^{12}$)}.}
\sm{We consider a concept-bottleneck pipeline that consists of a concept space $C$ that is about the attributes of the specific digit type.
We focus on the objective of 
\textit{interpretable classification} of the odd/even classes, so that} the mapping $X \rightarrow C$, with $C=Y^{(1:10)}$ is learned by a ResNet10 classifier $g$ \cite{he2016deep}  trained from scratch, while a LEN $f$ is used to learn both the mapping and the explanation as a function $C \rightarrow E$, with $E=Y^{(11:12)}$.


\paride{MNIST E/O (ICLU) - {\scriptsize CB, Interpretable clustering}.}
 The same dataset has been used with the objective of 
\textit{interpretable clustering}.  \sm{In this case, we considered a concept space $C = Y = [0,1]^{12}$} which comprise both the digits and the even/odd labels. The mapping $X \rightarrow C$ is learned again by a Resnet10 model $g$ trained as a multi-label classifier, while the LEN $f$ performs clustering from $C \rightarrow E$, where $E = [0,1]^{2}$ space, in order two extract two clusters \sm{that are expected to group data due to two properties of the concept space that are not-defined in advance.}

\paride{CUB - {\scriptsize CB, Interpretable Classification}.}
The Caltech-UCSD Birds-200-2011 dataset (CUB, \cite{wah2011caltech}) is a fine-grained classification dataset. It includes 11,788 images representing $200$ different bird species. \sm{Moreover,} 312 \sm{lower-level} binary attributes have been also attached to each image representing visual characteristics (color, pattern, shape) of particular parts (beak, wings, tail, etc.). Attributes annotations, however, are quite noisy. Similarly to \cite{koh2020concept}, we denoised attributes by considering class-level annotations, i.e. a certain attribute is set as present only if it is also present in at least 50\% of the images of the same class. Furthermore we only considered attributes present in at least 10 classes (species) after this refinement. \sm{In the end, a total of $108$ \sm{lower-level} attributes have been retained and paired with the higher-level attributes about the $200$ species, so that each image is represented with binary targets in $Y = [0,1]^{200+108=308}$ (where the first $108$ targets are for the lower-level attributes). In a concept-bottleneck pipeline with the LEN objective of \textit{interpretable classification}, 
the mapping $X \rightarrow C$ from images to lower-level concepts ($C=Y^{(1:108}$) is performed again with a ResNet10 model $g$ trained from scratch while a LEN $f$ learns the final function that classifies the  bird specie, $C \rightarrow E$, with $E=Y^{(109:308)}$.}

\paride{V-Dem - {\scriptsize Cascading, Interpretable classification.}}
Varieties of Democracy (V-Dem, \cite{pemstein2018v,coppedge2021v}) is a dataset containing a collection of indicators of latent regime characteristics over 202 countries from 1789 to 2020. The database includes $483$ low-level indicators (e.g. media bias, party ban, high-court independence, initiatives permitted, etc), $82$ mid-level indices (e.g. freedom of expression, freedom of association, equality before the law, etc), and 5 high-level indices of democracy principles (i.e. electoral, liberal, participatory, deliberative, and egalitarian). \sm{In the experiments, we considered each example to be paired with low/mid level indices and the information on electoral/non-electoral democracies taken from high level indices. Keeping the same order of the previous description, each sample is then paired with binary targets in $Y=[0,1]^{483+82+2=567}$. We considered the \textit{interpretable classification} objective in the problem of classifying electoral/non-electoral democracies in a cascading LEN with two LEN components, where $C_1=Y^{(1:483)}$, $E_1=C_2=Y^{(484:566)}$, and $E_2=Y^{(566:567)}$. In detail, cascading LENs are trained to learn the map $C_1 \rightarrow E_1$ with $E_1=C_2$ and $C_2 \rightarrow E_2$, with $E_2=Y^{(566:567)}$. We measured the quality of the rules extracted by the second LEN.}

\subsection{Metrics}
\label{sec:exp_metrics}
Seven metrics are used to compare the performance of LENs with respect to state-of-the-art approaches.
While measuring classification metrics is necessary for models to be viable in practice to perform interpretable classification tasks, assessing the quality of the explanations is required to justify their use for explainability. In contrast with other kinds of explanations, logic-based formulas can be evaluated quantitatively. Given a classification problem, first we extract a set of rules from a trained model and then we test/\sm{evaluate} each explanation on an unseen set of samples.
The results for each metric are reported in terms of the mean and standard deviation, computed over a 10-fold cross validation \cite{krzywinski2013error}. Only in the CUB experiments a 5-fold cross validation is performed due to timing issues related to BRL (each fold required about 3 hours)---\sm{competitors were described in Section~\ref{sec:related}.}
In particular, for each experiment we consider the following metrics.

\begin{figure}[!t]
    \centering
    \includegraphics[width=.9\textwidth,trim=0 15 0 0,clip]{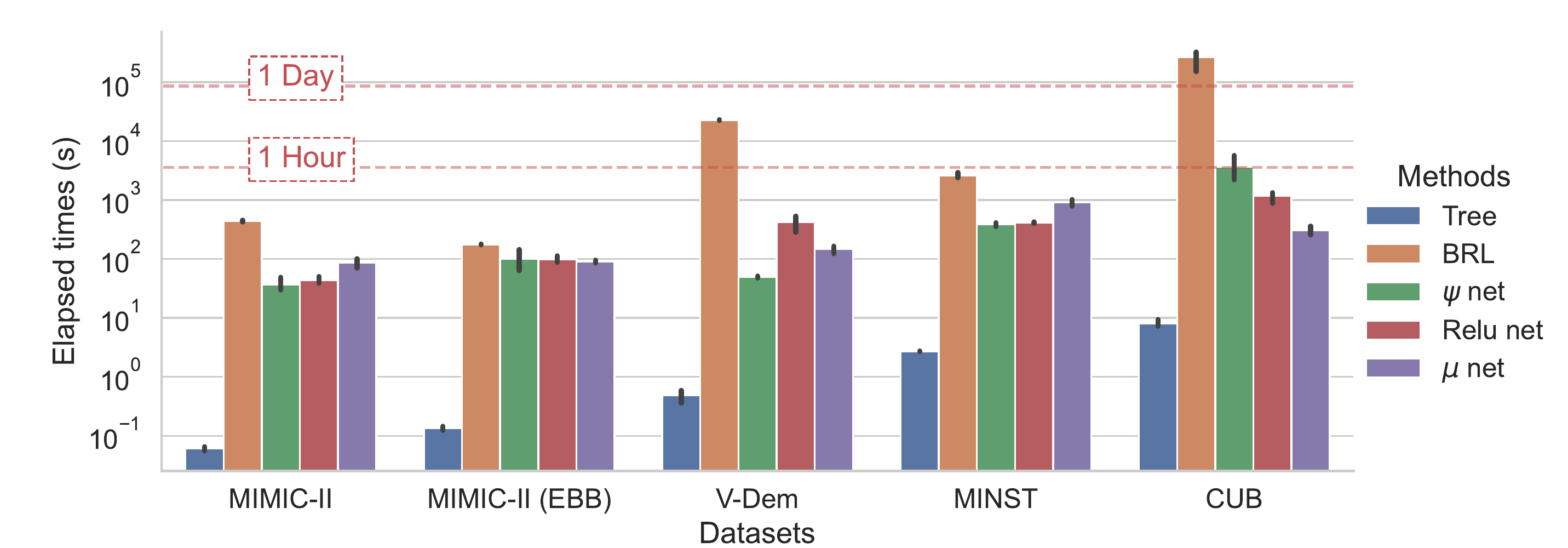}
    \caption{Rule extraction time (seconds). Time required to train models and to extract the explanations. 
    Error bars show the 95\% mean confidence interval.}
    \label{fig:time}
\end{figure}

\begin{itemize}
\item \textit{Model accuracy}: it measures how well \sm{the  LEN or the competitor classifier correctly identifies the target classes, in the case of interpretable classification.} When the LEN explains the predictions of a black-box classifier, this metric represents the accuracy of the model in mimicking the black-box classifier (Table~\ref{tab:model-accuracy}). 
\item \textit{Explanation accuracy}: it measures how well the extracted logic formula correctly identifies the target class  (Table~\ref{tab:explanation-accuracy}).
\item \textit{Complexity of an explanation}: it measures how hard would it be for a human being to understand the logic formula (Table~\ref{tab:complexity}). \sm{This is simulated} by standardizing the explanations in disjunctive normal form and then by counting the number of terms of the standardized formula.
\item \textit{Fidelity of an explanation}: it measures how well the predictions obtained by applying the extracted explanations match the predictions obtained when simply  using the classifier (Table~\ref{tab:fidelity}). When the LEN is the classifier itself (i.e. interpretable classification), this metric represents the match between the extracted explanation and the LEN prediction. Instead, when the LEN is explaining the predictions of a black-box classifier, this metric represents the \sm{agreement between the extracted explanation and the prediction of black-box classifier.}
\item \textit{Rule extraction \sm{time}}: it measures the time required to obtain an explanation from scratch (Fig.~\ref{fig:time}). It is computed as the sum of the time required to train the model and the time required to extract the formula from a trained model. This is justified by the fact that for some models, like BRL, training and rule extraction consist of just one simultaneous process.
\item \textit{Consistency of an explanation}: it measures the similarity of the extracted explanations over different runs (Table~\ref{tab:consistency}). It is computed by counting how many times the same concepts appear in the logic formulas over different folds of a 5-fold cross-validation or over 5 different initialization seeds. 

\end{itemize}

\begin{table}[!h]
\centering
\caption{Model accuracy (\%). The two best results are in bold.}
\label{tab:model-accuracy}
\resizebox{0.92\textwidth}{!}{\begin{tabular}{lll|lll}
\toprule
{} &              Tree &               BRL &        $\psi$ net &          ReLU net &         $\mu$ net \\
\midrule
\textbf{MIMIC-II} &  $77.53 \pm 1.45$ &  $76.40 \pm 1.22$ &  $77.19 \pm 1.09$ &  $\bf{80.11} \pm 1.87$ &  $\bf 80.00 \pm 0.95$ \\
\textbf{vDem    } &  $85.61 \pm 0.57$ &  $\bf 91.23 \pm 0.75$ &  $89.78 \pm 1.64$ &  $\bf{92.08} \pm 0.37$ &  $90.40 \pm 0.51$ \\
\textbf{MNIST E/O} &  $99.75 \pm 0.01$ &  $99.80 \pm 0.02$ &  $99.80 \pm 0.03$ &  $\bf{99.88} \pm 0.02$ &  $\bf 99.83 \pm 0.01$ \\
\textbf{CUB     } &  $81.62 \pm 1.17$ &  $90.79 \pm 0.34$ &  $91.92 \pm 0.27$ &  $\bf{92.29} \pm 0.40$ &  $\bf 92.21 \pm 0.33$ \\
\textbf{MIMIC-II (EBB)} &  $77.53 \pm 1.45$ &  $77.87 \pm 1.24$ &  $76.74 \pm 1.52$ &  $\bf{80.00} \pm 0.95$ &  $\bf{79.44} \pm 0.97$ \\
\bottomrule
\end{tabular}}
\end{table}

\begin{table}[!h]
\centering
\caption{Explanation accuracy (\%). The two best results are in bold (in case of ties in the average values, we highlighted all the involved models).}
\label{tab:explanation-accuracy}
\resizebox{0.97\textwidth}{!}{\begin{tabular}{lll|lll}
\toprule
{} &              Tree &               BRL &        $\psi$ net &          ReLU net &         $\mu$ net \\
\midrule
\textbf{MIMIC-II} &  $69.15 \pm 2.24$ &  $\bf 70.59 \pm 2.17$ &  $49.51 \pm 3.92$ &  $70.28 \pm 1.67$ &  ${\bf 71.84 \pm 1.82}$ \\
\textbf{vDem    } &  $85.45 \pm 0.58$ &  ${\bf 91.21 \pm 0.75}$ &  $67.08 \pm 9.68$ &  $\bf 90.21 \pm 0.55$ &  $88.18 \pm 1.07$ \\
\textbf{MNIST E/O} &  $\bf 99.74 \pm 0.01$ &  ${\bf 99.79 \pm 0.02}$ &  $65.64 \pm 5.05$ &  $\bf 99.74 \pm 0.02$ &  $\bf 99.74 \pm 0.02$ \\
\textbf{CUB     } &  $89.36 \pm 0.92$ &  ${\bf 96.02 \pm 0.17}$ &  $76.10 \pm 0.56$ &  $87.96 \pm 2.81$ &  $\bf 93.69 \pm 0.27$ \\
\textbf{MIMIC-II (EBB)} &  $69.15 \pm 2.24$ &  $\bf{71.68} \pm 2.21$ &  $51.71 \pm 4.78$ &  $70.53 \pm 1.56$ &  $\bf{71.84} \pm 1.82$ \\
\bottomrule
\end{tabular}}
\end{table}

\begin{table}[!h]
\centering
\caption{Complexity of explanations. The two best results are in bold (the lower, the better).}
\label{tab:complexity}
\resizebox{1.0\textwidth}{!}{\begin{tabular}{lll|lll}
\toprule
{} &              Tree &                   BRL &         $\psi$ net &           ReLU net &         $\mu$ net \\
\midrule
\textbf{MIMIC-II} &  $66.60 \pm 1.45$ &     $57.70 \pm 35.58$ &   $\bf 20.60 \pm 5.36$ &  $39.50 \pm 11.62$ &  ${\bf 15.80 \pm 1.37}$ \\
\textbf{vDem    } &  $30.20 \pm 1.20$ &    $145.70 \pm 57.93$ &    ${\bf 5.40 \pm 2.70}$ &   $18.40 \pm 2.17$ &   $\bf 9.10 \pm 0.94$ \\
\textbf{MNIST E/O} &  ${\bf 47.50 \pm 0.72}$ &  $1352.30 \pm 292.62$ &  $96.90 \pm 10.01$ &   $\bf 73.30 \pm 5.77$ &  $80.50 \pm 4.85$ \\
\textbf{CUB     } &  $45.92 \pm 1.16$ &       ${\bf 8.87 \pm 0.11}$ &   $15.96 \pm 0.96$ &   $60.57 \pm 6.95$ &  $\bf 14.65 \pm 0.16$ \\
\textbf{MIMIC-II (EBB)} &  $66.60 \pm 1.45$ &     $40.50 \pm 32.46$ &   $\bf{24.30} \pm 5.40$ &  $61.30 \pm 13.61$ &  $\bf{15.80} \pm 1.37$ \\
\bottomrule
\end{tabular}}
\end{table}

\begin{table}[!h]
\centering
\caption{Fidelity of explanations (\%). Tree and BRL trivially get 100\%. We highlighted in bold the best LEN model.}
\label{tab:fidelity}
\resizebox{0.95\textwidth}{!}{\begin{tabular}{lll|lll}
\toprule
{} &               Tree &                BRL &         $\psi$ net &          ReLU net &         $\mu$ net \\
\midrule
\textbf{MIMIC-II} &  $100.00 \pm 0.00$ &  $100.00 \pm 0.00$ &   $51.63 \pm 6.68$ &  $75.62 \pm 2.07$ &  $\bf 88.37 \pm 2.73$ \\
\textbf{vDem    } &  $100.00 \pm 0.00$ &  $100.00 \pm 0.00$ &  $69.67 \pm 10.43$ &  $\bf 96.36 \pm 0.64$ &  $94.73 \pm 2.16$ \\
\textbf{MNIST E/O} &  $100.00 \pm 0.00$ &  $100.00 \pm 0.00$ &   $65.68 \pm 5.05$ &  $\bf 99.85 \pm 0.02$ &  $99.83 \pm 0.02$ \\
\textbf{CUB     } &  $100.00 \pm 0.00$ &  $100.00 \pm 0.00$ &   $77.34 \pm 0.52$ &  $89.28 \pm 2.90$ &  $\bf 95.21 \pm 0.25$ \\
\textbf{MIMIC-II (EBB)} &  $100.00 \pm 0.00$ &  $100.00 \pm 0.00$ &   $55.72 \pm 8.55$ &  $74.14 \pm 2.32$ &  $\bf{88.37} \pm 2.73$ \\
\bottomrule
\end{tabular}}
\end{table}

\begin{table}[!h]
\centering
\caption{Rule consistency (\%). The two best results are in bold.}
\label{tab:consistency}
\resizebox{0.65\textwidth}{!}{\begin{tabular}{lll|lll}
\toprule
{} &     Tree &       BRL & $\psi$ net &  ReLU net & $\mu$ net \\
\midrule
\textbf{MIMIC-II} &  $40.49$ &   $30.48$ &    $27.62$ &   $\bf{53.75}$ &   $\bf{71.43}$ \\
\textbf{vDem    } &  $\bf{72.00}$ &   $\bf{73.33}$ &    $38.00$ &   $64.62$ &   $41.67$ \\
\textbf{MNIST E/O} &  $41.67$ &  $\bf{100.00}$ &    $96.00$ &  $\bf{100.00}$ &  $\bf{100.00}$ \\
\textbf{CUB     } &  $21.47$ &   $42.86$ &    $41.43$ &   $\bf{44.17}$ &   $\bf{76.92}$ \\
\textbf{MIMIC-II (EBB)} &  $40.49$ &   $40.00$ &    $25.71$ &   $\bf{58.67}$ &   $\bf{71.43}$ \\
\bottomrule
\end{tabular}}
\end{table}

\subsection{Results and discussion}
\sm{Our results are organized in order to compare the behavior of the models in all the considered tasks jointly, with the exception of interpretable clustering, that will be discussed in a separate manner.}
Experiments of Table~\ref{tab:model-accuracy} show how \sm{all the considered out-of-the-box LENs} generalize better than decision trees on complex Boolean functions (e.g. CUB) as expected \cite{tavares2020understanding}\sm{, and they usually outperform BRL as well. LENs based on ReLU networks are definitely the ones with better classification accuracy, confirming the intuitions reported in Section~\ref{subsec:ReLU}. $\mu$ nets, however, are quite close in terms of classification accuracy.}
These results must be paired with the ones of Table~\ref{tab:explanation-accuracy}, since having high classification performance and very low explanation quality would represent a major issue. 
For most experiments the formulas extracted from LENs are either better or almost as accurate as the formulas found by decision trees or mined by BRL, \sm{even if the top performance are reached by BRL. What makes LENs formulas preferable with respect to BRL is the significantly reduced complexity of the considered explanations, as shown in Table~\ref{tab:complexity}. Notice how less complex rules implies more readable formulas, that is a crucial feature in the context of Explainable AI.} More specifically, the complexity of LENs explanations is usually lower than the complexity of the rules extracted both from a decision tree\footnote{Decision trees have been limited to a maximum of 5 decision levels in order to extract rules of comparable length with the other methods.} or mined by BRL. \sm{Comparing the results of the different out-of-the-box LENs, we observe that $\psi$ networks yield moderately complex explanations, sometimes with limited accuracy, while ReLUs, due to the variability of example-level rules, lead to more complex explanations. Overall, the case of $\mu$ networks is the most attractive one, confirming the quality of their parsimony/pruning criterion.}

\sm{Moving to more fine-grained details, Table~\ref{tab:fidelity} shows how white-box models, like decision trees and BRL, outperform LENs in terms of fidelity. This is due to the fact that such models make predictions based on the explanation directly. Therefore fidelity is (trivially)} always 100\%. However, we see how the fidelity of the formulas extracted by the $\mu$ and the ReLU network is often higher than $90\%$. This means that almost any prediction has been correctly explained, making these networks very close to white boxes.
Fig.~\ref{fig:time} compares the rule extraction times. BRL is the slowest rule extractor over all the experiments, and LENs are faster by one to three orders of magnitude. In two cases BRL takes about 1 hour to extract an explanation and over 1 day in the case of CUB, making it unsuitable for supporting decision making. Decision trees are the fastest overall. 
In terms of consistency, LENs seem to provide more stable results, but overall there is not a dominant method (see Table~\ref{tab:consistency}). 
We can see, instead, how this result is clearly impacted by the considered dataset/task. \sm{Our intuition is that those datasets that are more coherently represented by the data in the different folds are expected to lead to more consistent behaviors.}

\begin{figure}[!t]
    \centering
    \includegraphics[width=0.32\textwidth, trim = 12 10 12 10, clip]{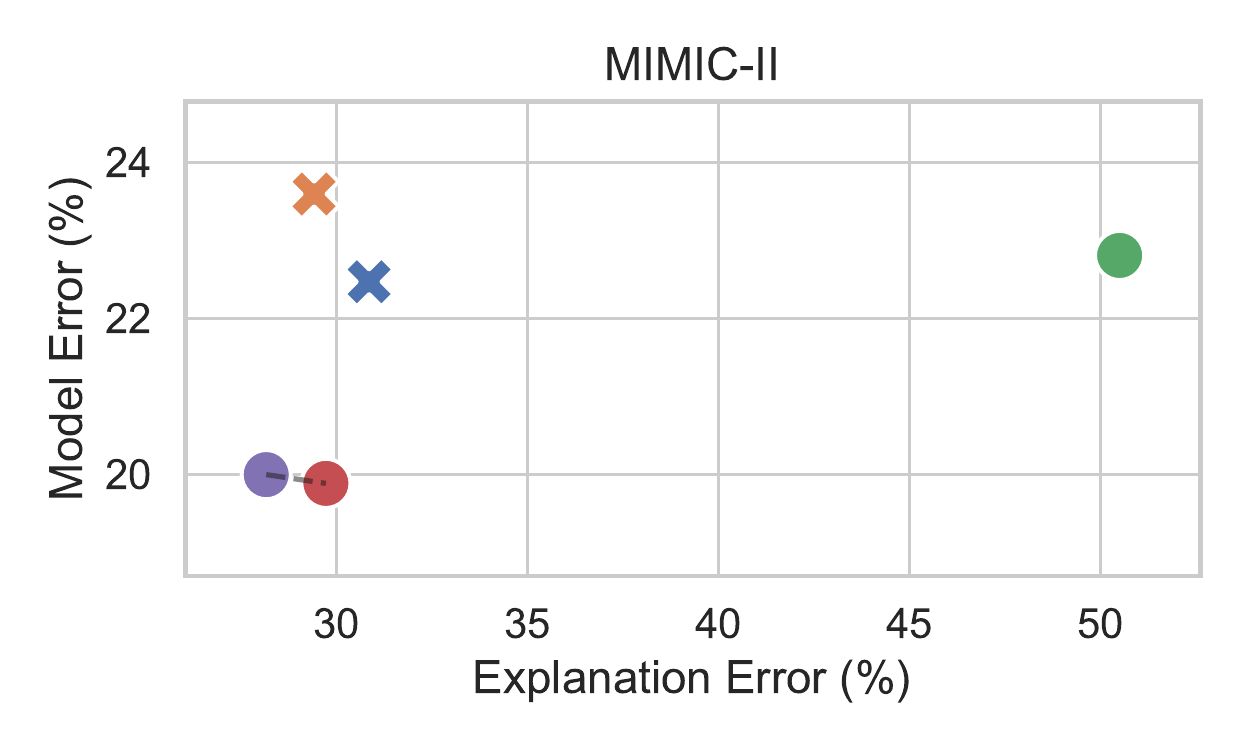}
    \includegraphics[width=0.32\textwidth, trim = 12 10 12 10, clip]{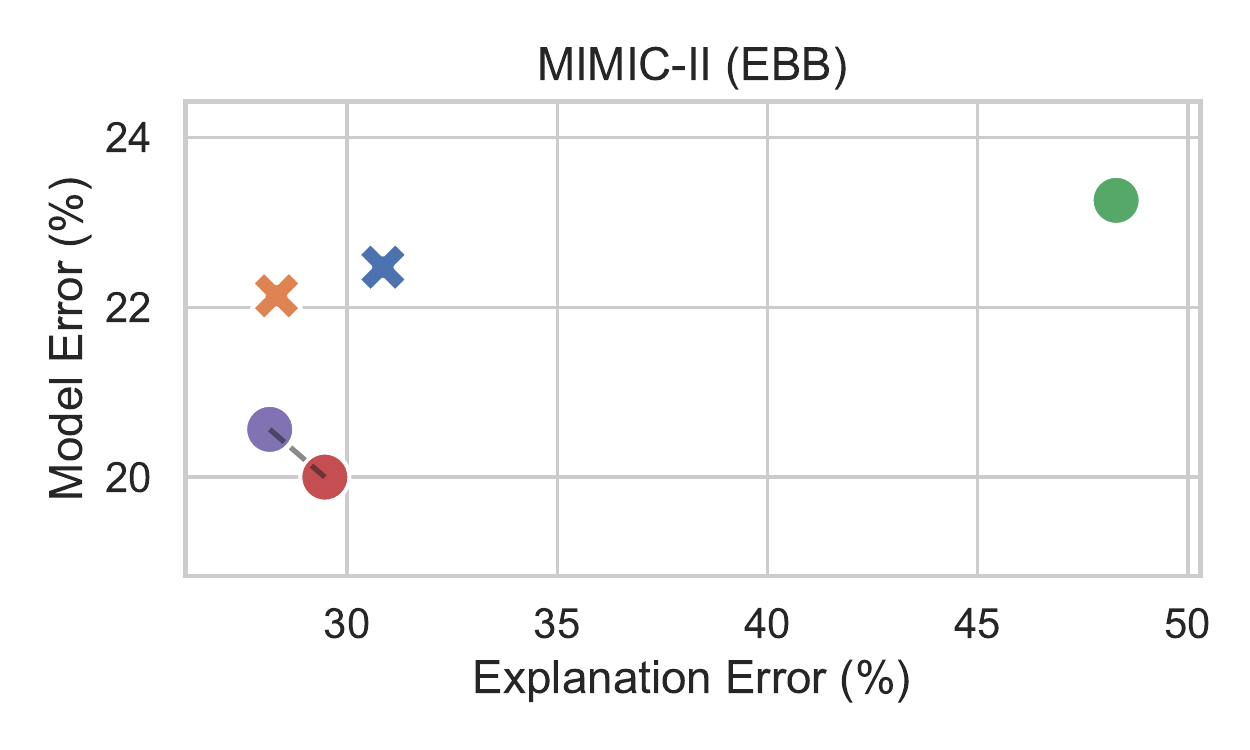}
    \includegraphics[width=0.32\textwidth, trim = 12 10 12 10, clip]{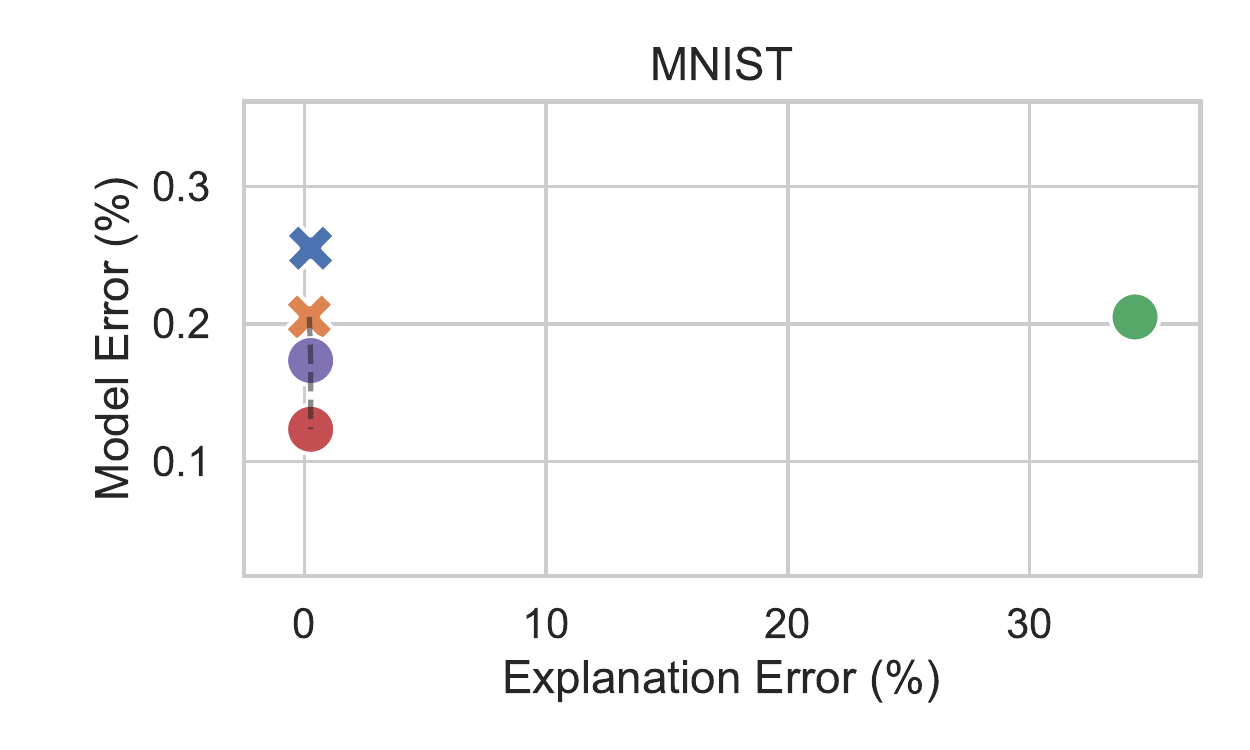}\\
    \includegraphics[width=0.32\textwidth, trim = 12 10 12 10, clip]{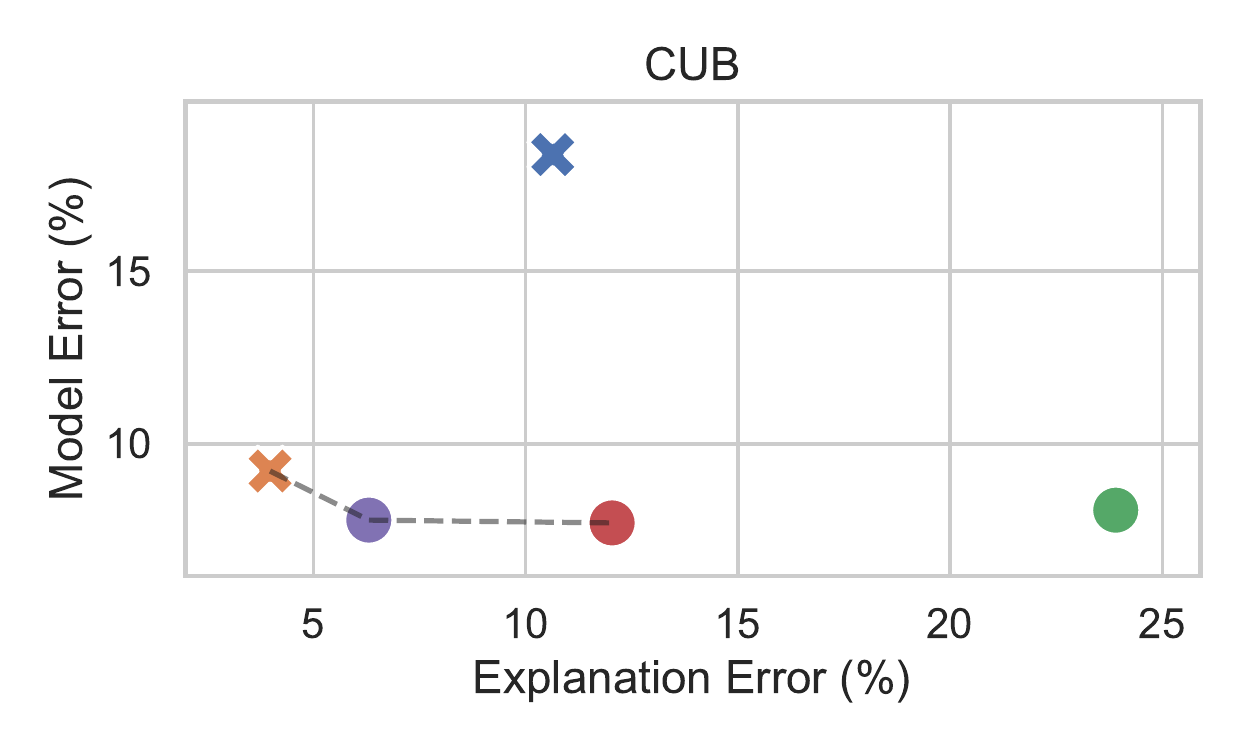}
    \includegraphics[width=0.32\textwidth, trim = 12 10 10 10, clip]{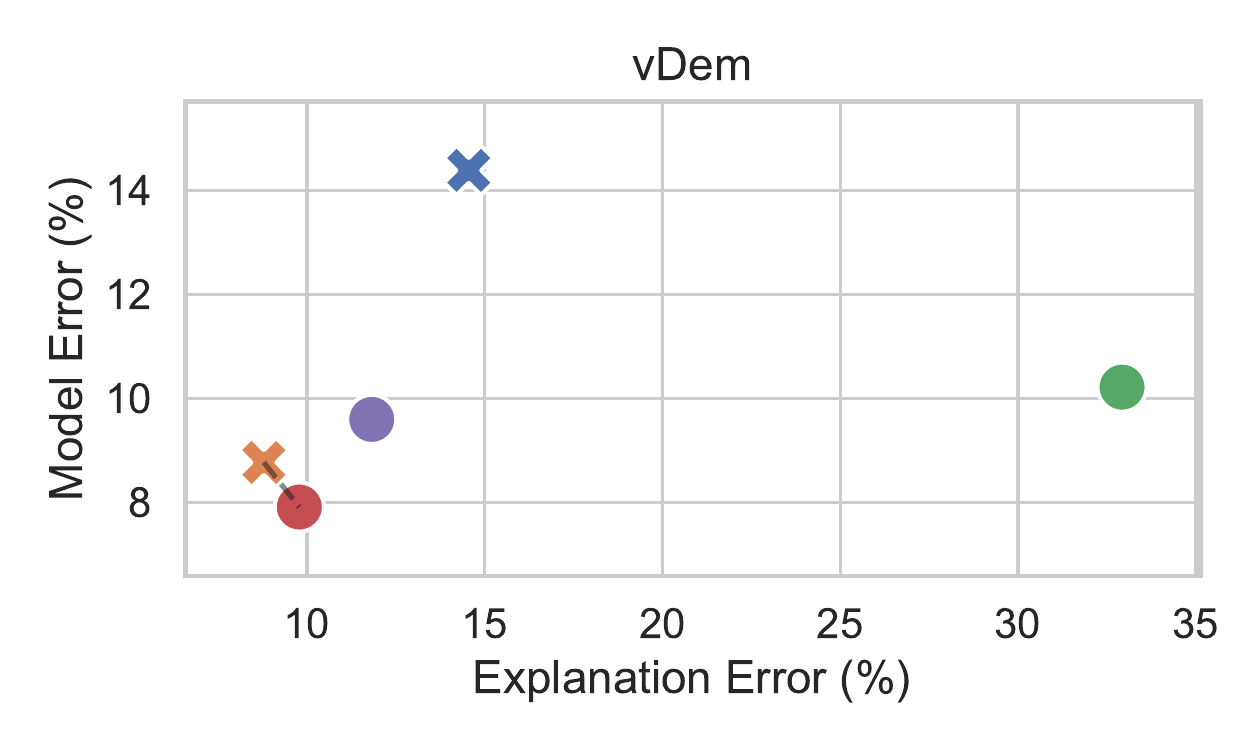}
    \\
    \includegraphics[width=0.7\textwidth]{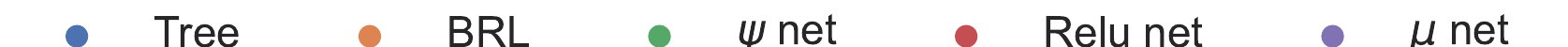}
    \\
    \caption{Pareto frontiers (dotted black line) in terms of average model test error and average explanation test error.} 
    \label{fig:multi-objective}
\end{figure}

Fig.~\ref{fig:multi-objective} shows \sm{a combined view of two of the main metrics considered so far, reporting} the Pareto frontiers \cite{marler2004survey} for each experiment in terms of the explanation and model error ($100$ minus the explanation/model accuracy). 
The figure provides a broad perspective showing how the LEN framework is flexible enough to instantiate different models having different compromises between explanation and classification accuracy. The limits of the $\psi$ network are overcome by the two other out-of-the-box LENs we investigated. In all considered tasks, the $\mu$ and ReLU LENs are always on the Pareto frontier, therefore providing, on average, the better accuracy vs. interpretability trade-offs.

\sm{Our experimental analysis quantitatively demonstrates the quality of LENs and of the FOL rules they extract. In order to provide the reader also a qualitative view on the type of rules that get extracted by the compared models, 
 Table~\ref{tab:best-rules} reports a representative subset of the formulas over the different tasks. 
 Comparing the LEN models, we can appreciate the compactness of rules coming from the $\mu$ and $\psi$ networks. 
 All the LEN-rules are way more compact than the ones from Trees, and usually also the ones from BRL.}
 
 {  \renewcommand{\arraystretch}{1}
\begin{table}[!ht]
\centering
\caption{A selection of some of the best rules extracted for each experiment. Each rule involves concept names that are taken from the respective dataset. We dropped the argument $(c)$ from each predicate to make the notation more compact, and each rule is intended to hold $\forall c \in C$. Rule from decision trees are not shown since they involve a very large number of terms.}
\vskip 2mm
\label{tab:best-rules}
\resizebox{\textwidth}{!}{%
\begin{tabular}[t]{ll|l}
\toprule
    & {Model}  & {Sample Rule} \\ 
\midrule
\multirow{6}{*}{\rotatebox{90}{\bf MIMIC-II}}     & $\mu$ net  & Death $\leftrightarrow$ stroke $\wedge$ age\_HIGH  $\wedge$ $\neg$atrial\_fibrillation   \\
                               & ReLU net   & Death $\leftrightarrow$ stroke $\wedge$ age\_HIGH $\wedge$ $\neg$atrial\_fibrillation $\wedge$ $\neg$sapsi\_first\_LOW $\wedge$ $\neg$sapsi\_first\_HIGH   \\
                               & $\psi$ net & Death $\leftrightarrow$ stroke $\vee$ bun\_first\_NORMAL $\vee$ sapsi\_first\_HIGH   \\
                               & Tree       & Death $\leftrightarrow$ $^*$formula is too long \\
                               & BRL        & Death $\leftrightarrow$ \makecell[tl]{(stroke $\wedge$ resp $\wedge$ $\neg$age\_LOW $\wedge$ $\neg$sapsi\_first\_LOW) \\
                               $\vee$ (stroke  $\wedge$ age\_HIGH $\wedge$ $\neg$age\_LOW  $\wedge$  $\neg$sapsi\_first\_LOW)}  \\ \hline
\multirow{6}{*}{\rotatebox{90}{\bf MIMIC-II {\tiny (EBB)}}} & $\mu$ net  & Death $\leftrightarrow$ atrial\_fibrillation $\wedge$ stroke $\wedge$ $\neg$sapsi\_first\_HIGH $\wedge$ $\neg$weight\_first\_NORMAL   \\
                               & ReLU net   & Death $\leftrightarrow$ stroke $\wedge$ $\neg$sapsi\_first\_LOW  $\wedge$  $\neg$sapsi\_first\_HIGH $\wedge$ $\neg$sodium\_first\_LOW   \\
                               & $\psi$ net & Death $\leftrightarrow$ stroke $\wedge$ age\_HIGH   \\
                               & Tree       & Death $\leftrightarrow$ $^*$formula is too long \\
                               & BRL        & Death $\leftrightarrow$ \makecell[tl]{stroke $\wedge$ age\_HIGH $\wedge$ weight\_first\_LOW $\wedge$ $\neg$sapsi\_first\_LOW) \\ $\vee$  (stroke $\wedge$ platelet\_first\_HIGH $\wedge$ weight\_first\_LOW $\wedge$ $\neg$sapsi\_first\_LOW }   \\  \hline
\multirow{17}{*}{\rotatebox{90}{\bf MNIST E/O}}         & $\mu$ net  & Even $\leftrightarrow$ $\neg$One $\wedge$ $\neg$Three $\wedge$ $\neg$Five $\wedge$ $\neg$Seven $\wedge$ $\neg$Nine   \\
                              & ReLU net   & Even $\leftrightarrow$ \makecell[tl]{(Zero $\wedge$ $\neg$One $\wedge$ $\neg$Two $\wedge$ $\neg$Three $\wedge$ $\neg$Four $\wedge$ $\neg$Five $\wedge$ $\neg$Six $\wedge$ $\neg$Seven $\wedge$ $\neg$Eight $\wedge$ $\neg$Nine) \\
                              $\vee$ (Two $\wedge$ $\neg$Zero $\wedge$ $\neg$One $\wedge$ $\neg$Three $\wedge$ $\neg$Four $\wedge$ $\neg$Five $\wedge$ $\neg$Six $\wedge$ $\neg$Seven $\wedge$ $\neg$Eight $\wedge$ $\neg$Nine) \\
                              $\vee$ (Four $\wedge$ $\neg$Zero $\wedge$ $\neg$One $\wedge$ $\neg$Two $\wedge$ $\neg$Three $\wedge$ $\neg$Five $\wedge$ $\neg$Six $\wedge$ $\neg$Seven $\wedge$ $\neg$Eight $\wedge$ $\neg$Nine) \\
                              $\vee$ (Six $\wedge$ $\neg$Zero $\wedge$ $\neg$One $\wedge$ $\neg$Two $\wedge$ $\neg$Three $\wedge$ $\neg$Four $\wedge$ $\neg$Five $\wedge$ $\neg$Seven $\wedge$ $\neg$Eight $\wedge$ $\neg$Nine) \\
                              $\vee$ (Eight $\wedge$ $\neg$Zero $\wedge$ $\neg$One $\wedge$ $\neg$Two $\wedge$ $\neg$Three $\wedge$ $\neg$Four $\wedge$ $\neg$Five $\wedge$ $\neg$Six $\wedge$ $\neg$Seven $\wedge$ $\neg$Nine) }  \\
                              & $\psi$ net & Even $\leftrightarrow$ \makecell[tl]{(Six $\wedge$ Zero $\wedge$ $\neg$One $\wedge$ $\neg$Seven)  $\vee$ (Six $\wedge$ Zero $\wedge$ $\neg$One $\wedge$ $\neg$Three)  \\
                              $\vee$ (Six $\wedge$ Zero $\wedge$ $\neg$Seven $\wedge$ $\neg$Three)  $\vee$ (Six $\wedge$ $\neg$One $\wedge$ $\neg$Seven $\wedge$ $\neg$Three) \\
                              $\vee$ (Zero $\wedge$ $\neg$One $\wedge$ $\neg$Seven $\wedge$ $\neg$Three)  $\vee$ (Six $\wedge$ Zero $\wedge$ $\neg$One $\wedge$ $\neg$Seven $\wedge$ $\neg$Three)}  \\
                              & Tree       & Even $\leftrightarrow$ $^*$formula is too long \\
                              & BRL        & Even $\leftrightarrow$  \makecell[tl]{(Six $\wedge$ $\neg$Three $\wedge$ $\neg$(Five $\wedge$ $\neg$Two)) $\vee$ (Eight $\wedge$ $\neg$Nine $\wedge$ $\neg$Six $\wedge$ $\neg$Three $\wedge$ $\neg$(Five $\wedge$ $\neg$Two))
                              \\ 
                              $\vee$ (Four $\wedge$ $\neg$Nine $\wedge$ $\neg$Six $\wedge$ $\neg$Three $\wedge$ $\neg$(Eight $\wedge$ $\neg$Nine) $\wedge$ $\neg$(Five $\wedge$ $\neg$Two) \\ 
                              $\wedge$ $\neg$(Seven $\wedge$ $\neg$Two)) $\vee$ (Two $\wedge$ $\neg$One $\wedge$ $\neg$Six $\wedge$ $\neg$Three $\wedge$ $\neg$(Eight $\wedge$ $\neg$Nine) \\
                              $\wedge$ $\neg$(Five $\wedge$ $\neg$Two) $\wedge$ $\neg$(Four $\wedge$ $\neg$Nine) $\wedge$ $\neg$(Seven $\wedge$ $\neg$Nine) \\
                              $\wedge$ $\neg$(Seven $\wedge$ $\neg$Two)) $\vee$ (Four $\wedge$ $\neg$Six $\wedge$ $\neg$Three $\wedge$ $\neg$(Eight $\wedge$ $\neg$Nine) \\ $\wedge$  $\neg$(Five $\wedge$ $\neg$Two) $\wedge$ $\neg$(Four $\wedge$ $\neg$Nine) $\wedge$ $\neg$(Seven $\wedge$ $\neg$Nine) $\wedge$ $\neg$(Seven $\wedge$ $\neg$Two) $\wedge$ \\ 
                              $\neg$(Two $\wedge$ $\neg$One)) $\vee$ (Zero $\wedge$ $\neg$Four $\wedge$ $\neg$Nine $\wedge$ $\neg$Six $\wedge$ $\neg$Three $\wedge$ $\neg$(Eight $\wedge$ $\neg$Nine) $\wedge$ \\
                              $\neg$(Five $\wedge$ $\neg$Two) $\wedge$ $\neg$(Four $\wedge$ $\neg$Nine) $\wedge$ $\neg$(Nine $\wedge$ $\neg$Zero) $\wedge$ $\neg$(One $\wedge$ $\neg$Two) \\
                              $\wedge$ $\neg$(Seven $\wedge$ $\neg$Nine) $\wedge$ $\neg$(Seven $\wedge$ $\neg$Two) $\wedge$ $\neg$(Two $\wedge$ $\neg$One))
                              }  \\ \hline
\multirow{14}{*}{\rotatebox{90}{\bf CUB}}          & $\mu$ net  &  Black\_foot\_albatross $\leftrightarrow$ \makecell[tl]{bill\_shape\_hooked\_seabird $\wedge$ \ size\_medium $\wedge$ \ wing\_pattern\_solid \\
                              $\wedge$ $\neg$wing color\_black $\wedge$ $\neg$underparts\ color\_white $\wedge$ $\neg$upper\_tail\_color\_grey \\
                              $\wedge$ $\neg$breast\_color\_white $\wedge$ $\neg$throat\_color\_white $\wedge$ $\neg$tail\_pattern\_solid \\
                              $\wedge$ $\neg$crown\_color\_white}   \\
                              & ReLU net   &  Black\_foot\_albatross $\leftrightarrow$  \makecell[tl]{ size\_medium  $\wedge$ $\neg$bill\_shape\_allpurpose  \\
                              $\wedge$ $\neg$upperparts\_color\_black  $\wedge$ $\neg$head\_pattern\_plain  \\  $\wedge$ $\neg$under\_tail\_color\_black  $\wedge$ $\neg$nape\_color\_buff  \\
                              $\wedge$ $\neg$wing\_shape\_roundedwings  $\wedge$ $\neg$shape\_perchinglike  \\
                              $\wedge$ $\neg$leg\_color\_grey  $\wedge$ $\neg$leg\_color\_black  $\wedge$ $\neg$bill\_color\_grey  \\
                              $\wedge$ $\neg$bill\_color\_black  $\wedge$ $\neg$wing\_pattern\_multicolored }\\
                              & $\psi$ net &  Black\_foot\_albatross $\leftrightarrow$  \makecell[tl]{(bill\_shape\_hooked\ seabird $\wedge$ tail\_pattern\_solid $\wedge$ $\neg$underparts\_color\_white \\
                              $\wedge$ ($\neg$breast\ color\ white $\vee$ $\neg$ \ wing\_color\_grey)) }  \\
                              & Tree       &  Black\_foot\_albatross $\leftrightarrow$  $^*$formula is too long \\
                              & BRL        &  Black\_foot\_albatross $\leftrightarrow$  \makecell[tl]{(bill\_shape\_hooked\_seabird $\wedge$ forehead\ color\ blue $\wedge$  $\neg$bill\_color\_black \\
                              $\wedge$  $\neg$nape\_color\_white)
                              $\vee$ (bill\_shape\_hooked\_seabird $\wedge$  $\neg$bill\_color\_black \\ 
                              $\wedge$  $\neg$nape\_color\_white $\wedge$  $\neg$tail\_pattern\_solid)} \\
\hline
\multirow{5}{*}{\rotatebox{90}{\bf V-Dem}}         & $\mu$ net  & Elect\_Dem $\leftrightarrow$ v2x\_freexp\_altinf $\wedge$ v2x\_frassoc\_thick $\wedge$ v2xel\_frefair $\wedge$ v2x\_elecoff $\wedge$ v2xcl\_rol   \\
                               & ReLU net   & Elect\_Dem $\leftrightarrow$ v2x\_freexp\_altinf $\wedge$ v2xel\_frefair $\wedge$ v2x\_elecoff $\wedge$ v2xcl\_rol   \\
                               & $\psi$ net & Elect\_Dem $\leftrightarrow$ v2x\_frassoc\_thick $\wedge$ v2xeg\_eqaccess   \\
                               & Tree       & Elect\_Dem $\leftrightarrow$ $^*$formula is too long \\
                               & BRL        & Elect\_Dem $\leftrightarrow$ v2x\_freexp\_altinf $\wedge$ v2x\_frassoc\_thick $\wedge$ v2xel\_frefair $\wedge$ v2x\_elecoff $\wedge$ v2xcl\_rol   \\ 
\bottomrule
\end{tabular}%
}
\end{table}}

\sm{A separate investigation is dedicated to the interpretable clustering objective in MNIST E/O (ICLU). In this scenario, competitors are not involved as they only operate in a supervised manner, while LENs can also handle the unsupervised discovery of FOL formulas.}
LENs are instructed to \sm{learn} two clusters considering the data represented in the concept space where the digit identity and the property of being even/odd are encoded. \sm{In this setting, we are interested in evaluating whether LENs can discover that data can be grouped into two balanced clusters of even and odd digits, and to provide explanations of them (such as, \textit{odd and not  even} and \textit{even and not odd}). We remark that this task does not use any supervisions, and we recall that LENs do not know that odd/even are mutually exclusive properties, but they are just two input concepts out of 12.} In Table \ref{tab:exp_metrics_mnist_mi}, \sm{all the previously introduced metrics are reported for this task, where the model accuracy is computed considering how the system develops clusters that match to the even/odd classes.} 
\begin{table}[!h]
\centering
\caption{MNIST E/O (ICLU). All the metrics are reported.}
\label{tab:exp_metrics_mnist_mi}
\resizebox{\textwidth}{!}{
\begin{tabular}{llllll}
\toprule
     Method & \sm{Model Acc} & Exp. Acc. (\%) &       Complexity & Extr. Time (sec) & Consistency (\%) \\
\midrule
 $\psi$  net &     $99.90 \pm 0.01$ &           $99.90 \pm 0.01$ &  $2.00 \pm 0.00$ &     $0.33 \pm 0.15$ &             $100.00$ \\
 $\mu$ net &     $94.91 \pm 4.99$ &           $95.41 \pm 4.50$ &  $2.30 \pm 0.64$ &     $2.36 \pm 0.75$ &              $35.00$ \\
   ReLU net &     $96.87 \pm 3.04$ &           $95.97 \pm 3.94$ &  $2.15 \pm 0.39$ &     $4.15 \pm 1.62$ &              $45.00$ \\
\bottomrule
\end{tabular}
}
\end{table}
We can see that all methods are capable of reaching very high level of cluster accuracy, therefore correctly identifying the even and odd groups. \sm{By inspecting the extracted rules, we observed that while all the LENs correctly consider odd and even as predicates that participate to the FOL rules,  only the} $\psi$ network consistently explains the two clusters in terms of only such important labels (even and odd), as we can see from the complexity of the rules (that is higher than 2 for the $\mu$ network and ReLU net) and from the consistency of terms used in the explanation ($100\%$ for the $\psi$ network).
Other examples and applications of interpretable clustering have been performed in previous works employing the $\psi$ network only \cite{ciravegna2020constraint, ciravegna2020human}.

\draft{
\subsection{LENs as Adversarial Defense}\label{sec:LEN4adv_att}
In nowadays machine learning literature, there is a serious concerns about the vulnerability of several machine learning models, such as neural networks, to the so called \textit{adversarial examples}. These are input samples maliciously altered with a slight perturbation that may lead to wrong predictions, even if they do not look to be visually altered for a human (in the case of images). This kind of issue has received a lot of attention, and several work has been done to develop some techniques, \textit{adversarial defenses}, to prevent an AI model from fraudulent \textit{adversarial attacks} \cite{miller2020adversarial,ozdag2018adversarial}.
Recently, it has been shown how explicit domain knowledge, expressed by a set of FOL formulas, can be used to discriminate if a certain sample has to be considered as adversarial \cite{melacci2020domain}, especially in case of multi-label classification. 
However this approach requires that the logic rules are already available for the considered task, so that a domain expert is expected to collect them.}

Interestingly, LENs can be applied to a set of clean data to automatically learn a set of FOL rules that may capture the important properties of the target domain. Then, these rules can be used to detect adversarial samples as in the framework introduced in \cite{melacci2020domain}, since they are exactly of the same type of the LEN ones.

\begin{figure}[!t]
    \centering
    \includegraphics[width=.45\textwidth]{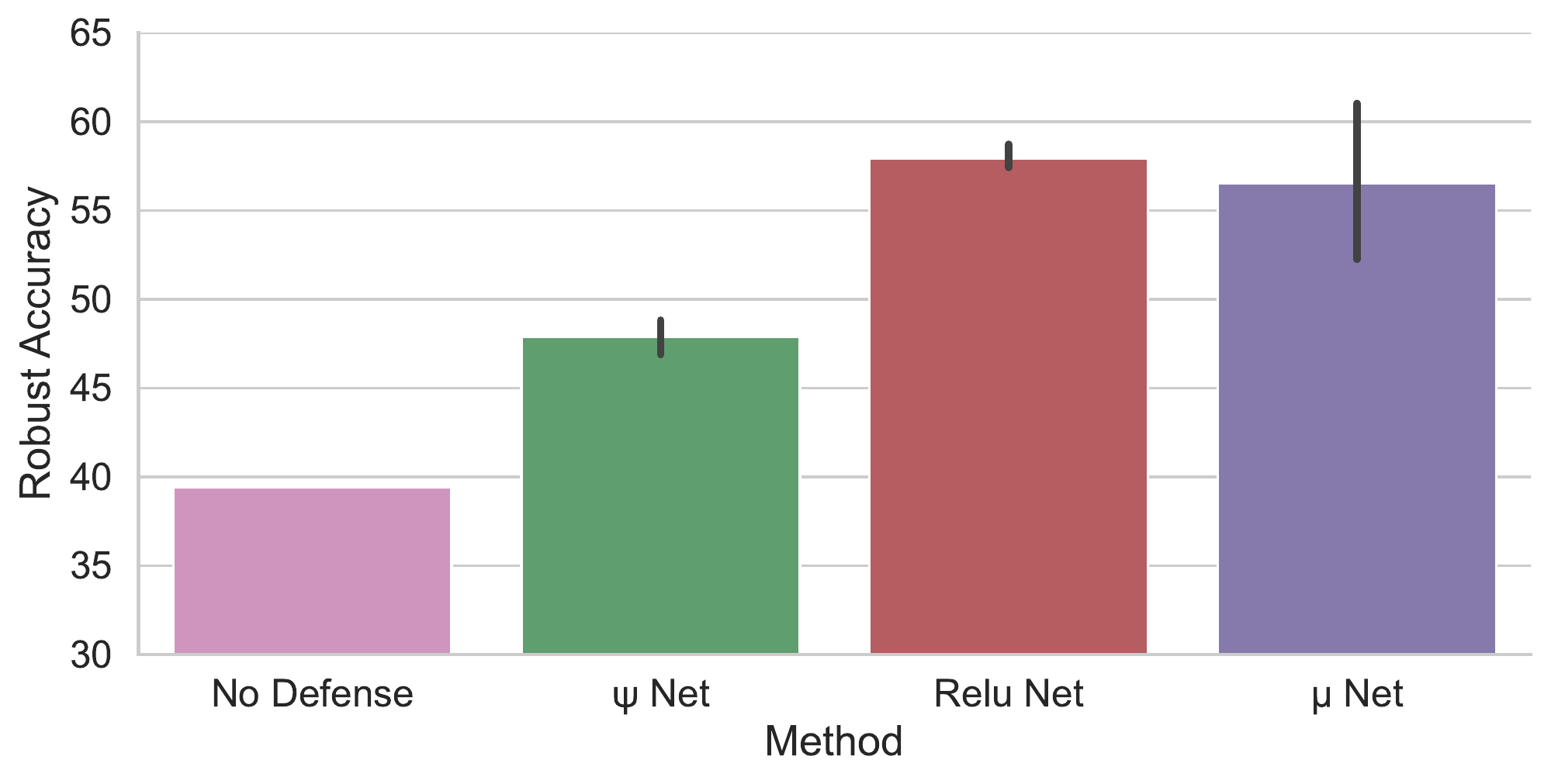}
    \caption{Quality of the explanations extracted by the LENs in terms of robust accuracy when used to defend a model \sm{from adversarial examples}. In pink, the accuracy of the model when facing a white box attack without any defense. Error bars show the 95\% confidence interval of the mean. }
    \label{fig:adv_def}
\end{figure}

\sm{Without any attempts of being exhaustive on this topic, we briefly explored whether what we stated in the previous sentence would work in practice. We considered the bird-identification dataset (CUB 200), facing the classification problem in its original multi-label fashion, where a convolutional neural network $g$ (ResNet18) is trained from scratch to classify both the 200 classes and the ones of the additional bird attributes. The FOL formulas automatically extracted by the LENs in the already discussed experimental experience are directly plugged in the rejection mechanism of \cite{melacci2020domain}, thus making $g$ equipped with a rejection function. We fixed the adversarial perturbation upper-bound to $\epsilon=0.5$, and we generated perturbed data attacking the 200 classes with the state-of-the-art attack APGD-CE \cite{croce2020reliable}, measuring the adversarial-aware accuracy described in \cite{melacci2020domain}, that here we refer to as robust accuracy.}
The experimental results for the three out-of-the-box LENs introduced in Section \ref{sec:outofthebox} are shown in Fig.~\ref{fig:adv_def}.
\sm{Interestingly, the accuracy of the attacked $g$ classifier increases in a significant manner using the rules extracted by LENs, confirming the validity of the proposed approach. The ReLU net leads to more detailed rules that better defend $g$, even if they introduce a larger computational burden in the defense mechanism, since they are usually more complex that the ones of the $\mu$ net.}

\section{Conclusions and Future Work} 
\label{sec:conclusion}

\sm{In this paper we presented a family of neural networks called Logic Explained Networks (LENs). We presented an in-depth study on the idea of using a neural model either to provide explanations for black-box or to solve a classification problem in an interpretable manner. Explanations are provided by First-Order Logic formulas, whose type is strongly interconnected with the learning criterion that drives LENs' training. Our framework covers a large set of computational pipelines and different user objectives. We investigated and experimented the case of three out-of-the-box LEN models, showing that they represent a balanced trade-off among a set of important properties, comparing favourably with state-of-the art white-box classifiers.}

The extraction of a first-order logic explanation requires symbolic input and output spaces. This constraint is the main limitation of our framework, as it narrows the range of applications down to symbolic input/output problems. In some contexts, such as computer vision, the use of LENs may require additional annotations and attribute labels to get a consistent symbolic layer of concepts. However, recent work may partially solve this issue leading to more cost-effective concept annotations \cite{ghorbani2019towards,kazhdan2020now}.
Another area to focus on might be the improvement of out-of-the-box LENs. The efficiency and the classification performances of fully interpretable LENs, i.e. $\psi$ network, is still quite limited due to the extreme pruning strategy. Even more flexible approaches, like $\mu$ networks, are not as efficient as standard neural models when tackling multi-class or multi-label classification problems, as they require a bigger architecture to provide a disentangled explanation for each class.
In our vision this framework would thrive and express its full potential by interacting with the external world and especially with humans in a dynamic way. In this case, logic formulas might be rewritten as sentences to allow for a more natural interaction with end users. Moreover, a dynamic interaction may call for an extended expressivity leading to higher-order or temporal logic explanations. Finally, in some contexts different neural architectures as graph neural networks might be worth exploring as they may be more suitable to solve the classification problem.

Current legislation in US and Europe binds AI to provide explanations especially when the economical, ethical, or financial impact is significant \cite{gdpr2017,law10code}. This work contributes to a lawful and safer adoption of some of the most powerful AI technologies allowing deep neural networks to have a greater impact on society. 
The formal language of logic provides clear and synthetic explanations, suitable for laypeople, managers, and in general for decision makers outside the AI research field.
The experiments show how this framework can be used to aid bias identification and to make black-boxes more robust to adversarial attacks. 
As out-of-the-box LENs are easy to use even for neophytes and can be effectively used to understand the behavior of an existing algorithm, our approach might be used to reverse engineer competitor products, to find vulnerabilities, or to improve system design. From a scientific perspective, formal knowledge distillation from state-of-the-art networks may enable scientific discoveries or confirmation of existing theories.

\section*{Acknowledgments}

We thank Ben Day, Dobrik Georgiev, Dmitry Kazhdan, and Alberto Tonda for useful feedback and suggestions.

This work was partially supported by the GO-DS21 project funded by the European Union’s Horizon 2020 research and innovation programme under grant agreement No 848077 and by the PRIN 2017 project RexLearn (Reliable and Explainable Adversarial Machine Learning), funded by the Italian Ministry of Education, University and Research (grant no. 2017TWNMH2). This work was also partially supported by TAILOR, a project funded by EU Horizon 2020 research and innovation programme under GA No 952215.

\bibliographystyle{apalike}  
\bibliography{references}  

\appendix

\section{Python APIs for LENs} \label{sec:apis}

In order to make LENs paradigms accessible to the whole community, we released "PyTorch, Explain!" \cite{barbiero2021lens}, a Python package\footnote{\url{https://pypi.org/project/torch-explain/}.} with an extensive documentation on methods and low-level APIs. 
Low levels APIs allow the design of custom LENs as illustrated in the example of Listing \ref{code:example2}.

\begin{center}
\begin{minipage}{0.7\textwidth}
\begin{lstlisting}[basicstyle=\ttfamily\tiny, language=Python, label=code:example2, caption=Example on how to use the "PyTorch Explain!" library to solve the XOR problem.]
import torch
from torch.nn.functional import one_hot
import torch_explain as te
from torch_explain.nn.functional import l1_loss
from torch_explain.logic.nn import psi
from torch_explain.logic.metrics import test_explanation, complexity

# train data
x_train = torch.tensor([
    [0, 0],
    [0, 1],
    [1, 0],
    [1, 1],
], dtype=torch.float)
y_train = torch.tensor([0, 1, 1, 0], dtype=torch.float).unsqueeze(1)

# instantiate a "psi network"
layers = [
    torch.nn.Linear(x_train.shape[1], 10),
    torch.nn.Sigmoid(),
    torch.nn.Linear(10, 5),
    torch.nn.Sigmoid(),
    torch.nn.Linear(5, 1),
    torch.nn.Sigmoid(),
]
model = torch.nn.Sequential(*layers)

# fit (and prune) the model
optimizer = torch.optim.AdamW(model.parameters(), lr=0.01)
loss_form = torch.nn.BCELoss()
model.train()
for epoch in range(6001):
    optimizer.zero_grad()
    y_pred = model(x_train)
    loss = loss_form(y_pred, y_train) + 0.000001 * l1_loss(model)
    loss.backward()
    optimizer.step()

    model = prune_equal_fanin(model, epoch, prune_epoch=1000, k=2)

# get first-order logic explanations for a specific target class
y1h = one_hot(y_train.squeeze().long())
explanation = psi.explain_class(model, x_train)

# compute explanation accuracy and complexity
accuracy, preds = test_explanation(explanation, x_train, y1h, target_class=1)
explanation_complexity = complexity(explanation)
\end{lstlisting}
\end{minipage}
\end{center}

\end{document}